%% file: main.tex
\documentclass[10pt,twocolumn,letterpaper]{article}

\usepackage{cvpr}              

\input{preamble}

\definecolor{cvprblue}{rgb}{0.21,0.49,0.74}
\usepackage[pagebackref,breaklinks,colorlinks,citecolor=cvprblue]{hyperref}
\usepackage{math}
\usepackage{multirow}
\usepackage{graphicx}

\title{CoE: Deep Coupled Embedding for Non-Rigid Point Cloud Correspondences}

\author{Huajian Zeng$^{1,2,*}$ 
\qquad
Maolin Gao$^{1,2,*}$
\qquad
Daniel Cremers$^{1,2}$\\
$^1$ Technical University of Munich 
\qquad
$^2$ Munich Center for Machine Learning
\qquad
$^*$ Equal contribution
}

\begin{document}

\twocolumn[{%
\renewcommand\twocolumn[1][]{#1}%
\maketitle
\vspace{-0.8cm}
\begin{center}%
    \centering%
    \captionsetup{type=figure}%
    \includegraphics[width=1\textwidth]{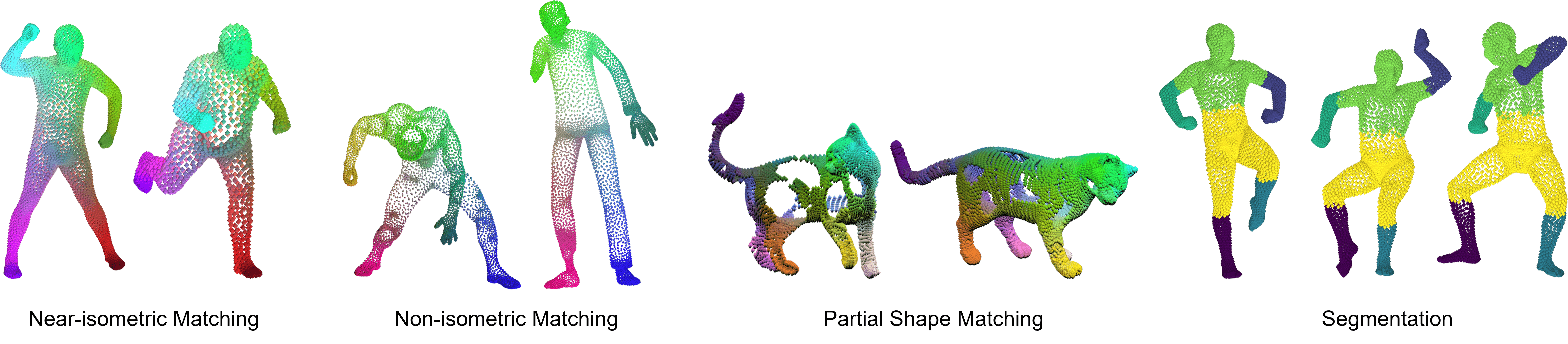}
    \caption{We propose a novel way to learn coupled embeddings of non-rigidly deformable shapes that are geometry-aware, robust and can be directly applied to retrieve accurate dense correspondences for near-isometric (\emph{left}), non-isometric (\emph{middle left}) and partial cases (\emph{middle right}). Furthermore, it can also be employed for other shape analysis tasks such as shape segmentation (\emph{right}).
    }
    \label{fig:teaser}
\end{center}%
}]

\makeatletter{\renewcommand*{\@makefnmark}{}
\footnotetext{The order of equally contributed authors can be changed freely.}\makeatother}

\makeatletter{\renewcommand*{\@makefnmark}{}
\footnotetext{Code: \url{https://github.com/zenghjian/coe}\makeatother}

\input{sec/0_abstract}    
\input{sec/1_introduction}

\input{sec/2_related_work}

\input{sec/3_background}

\input{sec/4_method}

\input{sec/5_experiment}
\input{sec/6_conclusion}

\newpage

{
    \small
    \bibliographystyle{ieeenat_fullname}
    \bibliography{main}
}
\input{sec/X_suppl}

\end{document}

%% file: preamble.tex
\usepackage[dvipsnames]{xcolor}


%% file: sec/0_abstract.tex
\begin{abstract}
The interest in matching non-rigidly deformed shapes represented as raw point clouds is rising due to the proliferation of low-cost 3D sensors. Yet, the task is challenging since point clouds are irregular and there is a lack of intrinsic shape information.
We propose to tackle these challenges by learning a new shape representation -- a per-point high dimensional embedding, in an embedding space where semantically similar points share similar embeddings. 
The learned embedding has multiple beneficial properties: it is aware of the underlying shape geometry and is robust to shape deformations and various shape artefacts, such as noise and partiality.
Consequently, this embedding can be directly employed to retrieve high-quality dense correspondences through a simple nearest neighbor search in the embedding space. 
Extensive experiments demonstrate new state-of-the-art results and robustness in numerous challenging non-rigid shape matching benchmarks and show its great potential in other shape analysis tasks, such as segmentation.
\end{abstract}

%% file: sec/1_introduction.tex
\section{Introduction}
Matching non-rigidly deformed 3D shapes is a long-standing and fundamental task in computer vision and graphics due to its ubiquitous role in many downstream tasks, such as shape editing, animation, medicine, statistical shape analysis, and robotics \cite{paravati2017animation, gainza2020nature, stegmann2002stat, sanchez2020slam}.
Often, 3D shapes are represented as (triangular) meshes, which consist of both points and their (intrinsic) neighborhood connectivities. However, with the proliferation of low-cost sensors, the interest in methods that can directly deal with raw point clouds is expanding rapidly.
Many (pointwise) shape descriptors have been proposed in the past decades, both hand-crafted \cite{sun2009concise, aubry2011wave, salti2014shot} and learned \cite{litany2017deep, attaiki2023clover, cao2023unsupervised}. Most of them are designed for shapes represented as triangular meshes and cannot be extended to point clouds without performance degradation \cite{marin2020lie, jiang2023nie, cao2023multimodal}. 
A particularly interesting type of descriptor is a (high-dimensional) embedding of shapes, which is a shape representation that is ideally invariant under natural deformations and, at the same time, contains enough information to perform geometry processing tasks.
Of particular interest is the global point signature (GPS) designed for triangular meshes \cite{rustamov2007gps}, which transforms the extrinsic coordinates of each surface point into a higher (potentially infinite) dimensional space by exploiting the scale and isometric invariance of eigenfunctions of the Laplace-Beltrami Operator (LBO). While being effective, it turns out to be unstable due to sign ambiguity and complex spectrum in the LBO eigen-decomposition. 
On the other hand, the seminal work by Ovsjanikov et al.~\cite{ovsjanikov2012functional} proposes to align these LBO spectral embeddings before searching for the correspondence in this embedding space. The alignment of the high-dimensional spectral embeddings is called functional maps (fmaps), which can be represented compactly as a low-dimensional matrix.
However, spectral embeddings are computed inefficiently by a non-differentiable eigen-decomposition of the LBO, which is sensitive to various practical artifacts, such as noise, partiality, and topological ``short circuits".

Inspired by \cite{kovnatsky2013coupled, eynard2015joint}, in which an (orthogonal) transformation of LBO eigenfunctions is estimated to obtain a consistent basis using manifold optimisation, we propose to leverage the power of deep learning to learn a coupled canonical embedding directly from raw point clouds,  which can recover the LBO eigenbasis as a special case.

Due to insights gained from the classical geometry processing, we can obtain high-quality dense correspondences directly via a simple proximity search in the embedding space by training a \emph{single} network, while all previous state-of-the-art methods have to train two networks \cite{marin2020lie, jiang2023nie}, underscoring the high practicability of our proposed method.
Furthermore, our learned embedding is aware of the underlying geometry of the surface, efficient to compute, robust to various shape artefacts and applicable for various shape analysis tasks such as correspondences and segmentation (c.f. Fig.~\ref{fig:teaser} \& Sec.~\ref{sec:exp}).
Extensive experiments show that our proposed method can robustly map extrinsic coordinates of shapes, which undergo various non-rigid deformations, to a canonical embedding space where corresponding points share similar embeddings.

\noindent In summary, our contributions are:
\begin{itemize}
    \item We propose a novel unsupervised way to learn per-point embeddings directly from raw point clouds under various non-rigid deformations. Inspired by classical geometry processing technique, our method is effective and simple that only requires to train a single network.
    \item In our learned embedding space, non-rigidly deformed shapes share similar and geometry-aware embeddings for corresponding points, which can be used for efficient matching by a simple nearest neighbor search.
    \item We show superior performance in a number of challenging non-rigid shape matching benchmarks, and unprecedented generalisation ability and robustness against different noise types, setting the new state-of-the-art.
    \item As a proof-of-concept, we show that our learned embeddings can be applied in other shape analysis tasks, such as partial shape matching and shape segmentation.
\end{itemize}

%% file: sec/2_related_work.tex
\section{Related Work}
The field of shape matching is vast and has rapidly developed over the past decades. Below we review the works which are most related to ours and can serve as baselines to our best knowledge. For a more comprehensive overview of the field, please refer to recent surveys \cite{sahillioglu2020survey, deng2022survey}.

\begin{figure}[t]
    \centering
    \includegraphics[width=0.47\textwidth]{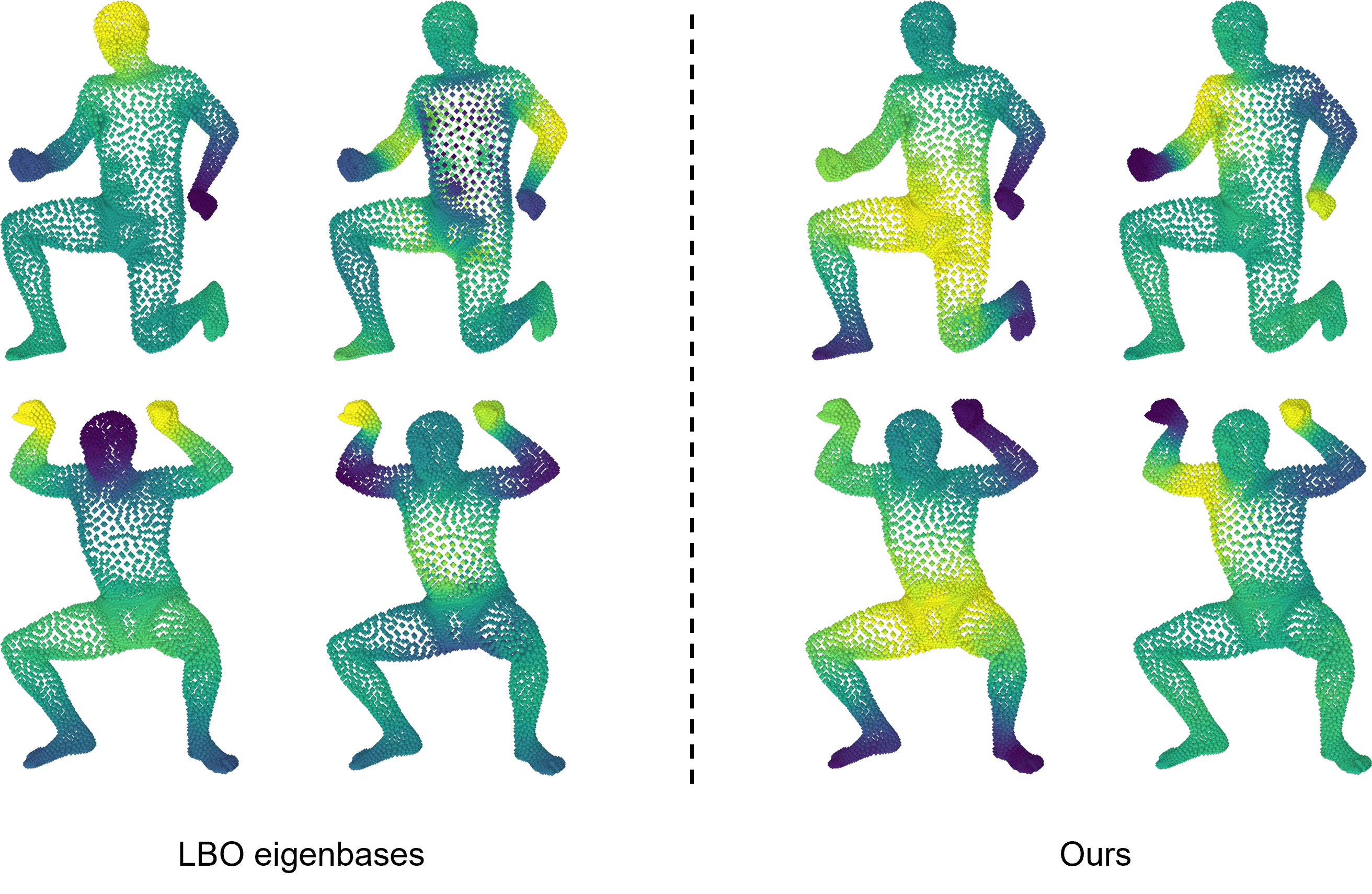}
        \caption{
        Examples of LBO eigenbases and our learned coupled embeddings on a pair of non-rigidly deformed shapes. Ours are consistent while LBO eigenbases suffer from sign flip (cf. Fig.~\ref{fig:supp_basis_m} for more examples).
        }
    \label{fig:embedding}
\end{figure}

\subsection{Pose Invariant Shape Representation}
The study of pose invariant shape representation dates back to the beautiful work by Torgerson in 1952 \cite{torgerson1952mds}, where he introduced a lower-dimensional embedding, which preserves the pairwise (geodesic) distances between all graph nodes as much as possible and finds applications in many tasks, such as visualisation and clustering.
This idea of dimensionality reduction has been further studied by \cite{coifman2005, reuter2006shapedna} based on the LBO. The work \cite{coifman2005} utilises the LBO eigenfunctions and shows that these spectral properties can be employed to embed the data into a Euclidean space based on a diffusion process.
Later approaches continue the exploration in the opposite direction, by embedding a surface into a higher dimensional space
\cite{rustamov2007gps,  reuter2010hierarchy, xia2022ge}. 
The GPS embedding \cite{rustamov2007gps} combines the LBO eigenvalues and eigenbases and in fact constructs a new surface in the infinite-dimensional space, which is invariant to (isometric) deformation. Impressive results on shape segmentation and clustering have demonstrated the effectiveness of the GPS embedding. 
From the standpoint of geodesic distance preservation, the authors of \cite{xia2022ge} design an embedding for the fast approximation of geodesics using a cascade strategy to gradually improve the accuracy of the approximation.

Most recently, deep-learning-based pose invariant embeddings are becoming prevalent. Most similar to our approach are the ones proposed in \cite{marin2020lie, jiang2023nie}.
DiffFmaps \cite{marin2020lie} proposes to learn a linearly invariant embedding from point clouds, which serves as a replacement for the pre-computed LBO eigenbasis in the fmaps framework. 
Improved robustness and better accuracy have been reported. 
Due to the employment of fmaps, it can efficiently regularise the maps in the functional space and incorporate structural regularisation. 
However, at the same time, it leads to the necessity of a \emph{second} separate network dedicated to feature learning, hence a more complex pipeline in practice. 
Moreover, it requires ground truth correspondences to train, which is eliminated in NIE \cite{jiang2023nie} by rigid pre-alignment (\ie~weak supervision), while retaining the dependency on a \emph{second} feature network. 
In contrast, we propose to look at the correspondences problem through the classical geometry processing lens and learn a canonical embedding of the shape, which can be used directly for finding correspondences. 
As we show later, our embedding is motivated by the LBO spectral embedding while remaining coupled across different non-rigidly deformed shapes.

\subsection{Basis Pursuit for Shape Analysis}
In another line of work \cite{kovnatsky2013coupled, eynard2015joint, huang2022multiway}, researchers start to ask the question: is it possible to obtain a better set of basis suitable for shape correspondences? 
Huang et al. \cite{huang2022multiway} proposes to learn a set of non-orthogonal bases and demonstrates its expressiveness and flexibility. However it internally converts the shape representation from point clouds to 3D voxel grid and requires careful engineering to obtain good results, such as post refinement and synchronisation. Moreover, it has to train a \emph{second} network dedicated to feature learning due to the employment of fmaps.
Kovnatsky et al. \cite{kovnatsky2013coupled} tries to approximately diagonalise the LBOs of two shapes simultaneously. To reduce the dimensionality of the solution space, it makes use of subspace parametrisation to compute an (orthonormally) transformed version of the LBO eigenbases. However, ground truth dense correspondences are required, which we ultimately would like to estimate, rendering this method less practical for shape matching tasks. 
This requirement has been relaxed in \cite{eynard2015joint} by requiring only sparse ground truth correspondences and dense descriptors. However its best performance still demands some ground truth labelling and both approaches involve complex manifold optimisation (cf. \cite{eynard2015joint} \& Tab.~\ref{tab:complete_shape}).
Inspired by \cite{kovnatsky2013coupled, eynard2015joint} together with our insights on pose invariant representations, we propose to learn coupled embeddings directly from data. As we show in Sec.~\ref{sec:method} \& \ref{sec:exp}, our proposed method can learn high-quality coupled embeddings from low-quality shape descriptors. This attributes to the careful design of our geometry-aware unsupervised loss and network architecture, which enables \emph{cross-communication} between shapes that is key for their coupling and consistency. 

\subsection{Learning on Point Clouds}
\label{subsec:rw_pcl_learning}
Point cloud is arguably the most common representation for 3D shapes. However, due to its irregularity (compared to e.g. volumetric grids and triangular meshes), learning directly on point clouds has only become possible in recent years, enabled by specially designed architectures such as \cite{qi2016pointnet, qi2017pointnetpp, wang2019dgcnn, sharp2022diffusionnet}. 
While there are many works to learn deep features on meshes,
tackling raw point clouds as the input 3D representation has been relatively less studied \cite{Groueix20183d, donati20geomfmap, sharma20weakly, huang2022multiway}. 
This is partially due to the missing intrinsic proximity information in point clouds, which can be very helpful in many geometry processing tasks, such as computing geodesic distances. 
However, incorrect or even inconsistent topology often complicates algorithms and makes it very challenging to recover from it \cite{eisenberger2023gmsm, cao2023unsupervised}. 
In this work, we choose raw point clouds as our 3D shape representation. In this sense, our method is most similar to \cite{donati20geomfmap, sharma20weakly}, which employ fmaps in designing their losses. 
While GeomFmap \cite{donati20geomfmap} is a supervised approach which requires ground truth pointwise correspondences, a more recent work \cite{sharma20weakly} shows that only an approximate pre-alignment of shapes can replace the costly demand in ground truth matches, which is often dubbed as weak supervision in the literature. 
We make use of the weak supervision same as in \cite{sharma20weakly}, since most datasets come already approximately rigidly aligned or can be easily aligned with very little manual intervention.

%% file: sec/3_background.tex
\section{Background and Notation}
\label{sec:background}
 
In this section, we briefly review coupled diagonalisation for a pair of input shapes and introduce our notations (Tab.~\ref{tab:notation}). See \cite{kovnatsky2013coupled,eynard2015joint} for a comprehensive discussion and the supplementary for an introduction of the LBO.

Given shapes $\cS$ and $\cT$ and their LBOs represented in stiffness matrices $\bLS$, $\bLT$ and mass matrices $\bMS$, $\bMT$, the coupled diagonalisation problem can be modelled as:
\begin{align}
    \label{eq:coupledDiag}
    \min_{\{\bPsi_i\}}  & \sum_{i\in\{\cS,\cT\}} \text{off}(\bPsitop_i \bL_i \bPsi_i) + \mu_c \|\bDS^\top \bMS \bPsiS - \bDT^\top \bMT \bPsiT\| \\
    &\text{s.t.}~~~~\bPsitop_i \bM_i \bPsi_i = \bI,~~\text{for}~~i\in\{\cS,\cT\} \nonumber
\end{align}

where $\bDS$ and $\bDT$ are given descriptors (f.e. ground truth correspondences as indicator functions) and $\bI$ is the identity matrix. 

The $\text{off}(\cdot)$ term ensures that the coupled bases behave as approximate LBO eigenbases by penalising the off-diagonal entries and 
we chose $\text{off}(\bPsitop \bL \bPsi) = \text{off}(\bPsitop \bL \bPsi; \eval)=\|\bPsitop \bL \bPsi - \eval\|$ 
throughout our experiments consistently, where $\eval$ is a diagonal matrix of eigenvalues of the respective LBO. This choice helps to select leading bases (bases corresponding to small eigenvalues) with increasing frequency, which are the most informative ones in shape matching \cite{ovsjanikov2012functional}.
The second term is a coupling term that encourages the corresponding descriptors to behave similarly in the respective bases, which amounts to coupling the bases and making them to ``speak the same language". 
Note that the basis of a 2-manifold is a generalisation of the (1D/2D) Fourier basis in the Euclidean space, which is fixed and always consistent.

\input{tables/table_notation}
The corresponding descriptors can be indicator (delta) functions representing (dense/sparse) ground truth pointwise correspondences, blobs or stable regions, distance functions and dense descriptors as discussed in~\cite{kovnatsky2013coupled, eynard2015joint}, and in practice some amount of ground truth information is required  for good performance~\cite{kovnatsky2013coupled, eynard2015joint} (also see Sec.~\ref{sec:exp}, Tab.~\ref{tab:complete_shape}), since the quality of the estimated coupled bases is strongly tied to the quality of the corresponding descriptors. 
Note that when $\mu_c \rightarrow 0$, problem \eqref{eq:coupledDiag} becomes separable and amounts to solving the LBO eigen-decomposition of $\cS$ and $\cT$ separately. 

Since Eq.~\eqref{eq:coupledDiag} does not scale well with the size of the shape, it makes the optimisation problem very challenging or even intractable for high resolution shapes. Therefore, the authors propose to solve a surrogate problem by subspace parameterisation, namely representing the coupled basis $\bPsi$ as a linear combination of the LBO eigenbasis $\bPhi$, \ie $\bPsi = \bPhi \bR$, where $\bR$ is a Stiefel matrix. 
Compared to the original problem in Eq.~\eqref{eq:coupledDiag}, this modification greatly reduced the computational complexity, however it 
still involves difficult manifold optimisation for only approximately solving the original one. 
Furthermore, it demands at least a sparse set of ground truth correspondences to obtain good coupled bases, which, unfortunately, makes it dependent on either sparse shape matching methods or manual labeling (to produce the sparse correspondences).

To overcome these issues, we propose to directly learn coupled embeddings without any ground truth correspondences and without any subspace parameterisation. As demonstrated below, we only require noisy easy-to-obtain pointwise feature descriptors, f.e.~heat kernel signature (HKS) \cite{sun2009concise}, out of which our network can learn high-quality embeddings which are coupled and can be used directly for shape correspondence tasks.

%% file: tables/table_notation.tex
\begin{table}[thb!]
    \centering
    \footnotesize
    \begin{tabular}{lp{4.5cm}}
        \toprule
        \textbf{Symbol} & \textbf{Description} \\
        \midrule
        $\cS $ & Source shape (point cloud)\\
        $\bVS\in \mathbb{R}^{n_\cS \times 3}$ & All $n_{\cS}$ points  of shape $\cS$ \\  
        $\bL_{\cS} \in \BR^{n_{\cS}\times n_{\cS}}$ & Stiffness matrix of shape $\cS$ \\   
        $\bM_{\cS} \in \BR^{n_{\cS}\times n_{\cS}}$ & Mass matrix of shape $\cS$ \\
        $\bD_{\cS} \in \BR^{n_{\cS} \times d}$ & Pointwise descriptors of shape $\cS$\\
        $\bPhiS\in \BR^{n_{\cS} \times k}$ & LBO Eigenfunctions of shape $\cS$\\
        $\eval_{\cS}\in \BR^{k \times k}$ & LBO Eigenvalues of shape $\cS$\\
        $\hat\bPsi_{\cS}\in \BR^{n_{\cS} \times k}$ & Intermediate embedding of shape $\cS$ \\
        $\bPsiS\in \BR^{n_{\cS} \times k}$ & Predicted embedding of shape $\cS$\\
        $\cT$ & Target shape (point cloud) \\
        \vdots & (analogous as above for $\cT$)\\
        $f_{\theta}$ & Embedding extractor with learnable $\theta$ \\
        $h_{\varphi}$ & Cross attention block with learnable $\varphi$ \\
        $\Pi_{\cS\cT}\in \BR^{n_{\cS} \times n_{\cT}}$ & Binary matching matrix from $\cS$ to $\cT$ \\ 
        \bottomrule
    \end{tabular}
    \caption{Summary of our notation used in the paper.}

    \label{tab:notation}
\end{table}

%% file: sec/4_method.tex
\section{Deep Coupled Embeddings}
\label{sec:method}

\begin{figure}[t!]
\hspace*{-1.2cm}
    \centering
    \includegraphics[width=0.65\textwidth]{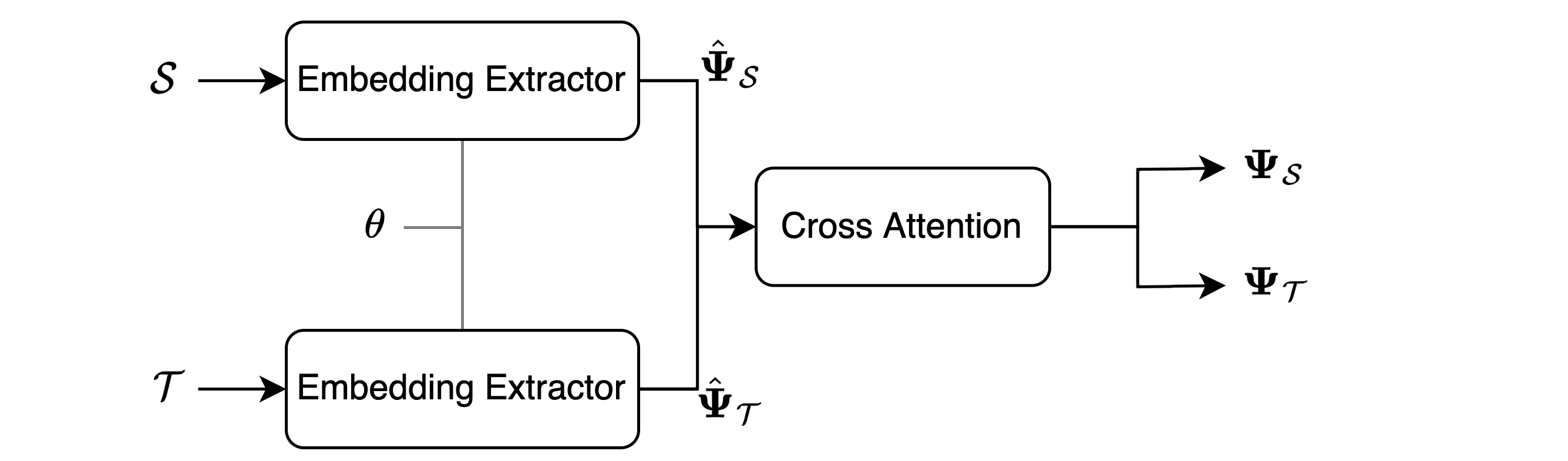}
        \caption{Pipeline overview. Given a pair of shapes $\cS$ and $\cT$ represented in point clouds,  Our embedding extractor -- ASAP DiffusionNet with shared weights $\theta$  (not to be confused with generative diffusion models~\cite{ho2020ddpm, song2021scorebased}), extracts the intermediate per-point embeddings $\hat\bPsi_\cS$ and $\hat\bPsi_\cT$, which are further refined by the subsequent cross attention block to output the final coupled embeddings $\bPsi_\cS$ and $\bPsi_\cT$.
        The cross attention block constructs a complete bipartite graph that connects every point on the shape $\cS$ with every points on the shape $\cT$ to enable their cross-communication.
        Our unsupervised loss encourages the predicted embeddings of both shapes to be coupled while closely resembling the LBO eigenbases.
        }
    \label{fig:pipeline}
\end{figure}

Real world shapes such as human and animals are intrinsically 2-dimensional compact manifolds and often embeded into the 3-dimensional Euclidean space and discretised as point clouds. 
It is of great interest to study the structure of the 2-manifold, rather than its Euclidean 3D embedding. One reason is that the intrinsic information of a shape (f.e. proximity) is ``hidden" in point clouds, despite its convenience to store and render, limiting its usage in shape geometry and analysis tasks.
Our network is designed primarily to recover the intrinsic proximity information, which can be used for direct retrieval of dense shape correspondences.

Given shapes $\cS$ and $\cT$ represented as point clouds $\bVS$ and $\bVT$, our proposed method learns their high-dimensional deep coupled embeddings $\bPsiS$ and $\bPsiT$, based on which accurate dense correspondences can be obtained via a simple proximity search in the embedding space.
In this section, We first introduce our network architecture design in Sec.~\ref{subsec:method_network}, which combines a recent variant of DiffusionNet \cite{sharp2022diffusionnet, attaiki2023clover} with cross attention to encourage information exchange during the learning process. 
Subsequently, we introduce our loss in Sec.~\ref{subsection:method_loss}. 
Note that our loss does not require any ground truth, hence enabling 3D representation learning in a fully data driven fashion.  
In Sec.~\ref{subsec:method_matching}, we explain our dense correspondence retrieval based on the learned embeddings. 
Throughout this section, we discuss the key design insight to achieve the coupling of learned embeddings and shed light to their geometric properties that are valuable for shape analysis tasks.

\subsection{Network Architecture}
\label{subsec:method_network}

Our network architecture is simple, efficient and comprises two main building blocks: an embedding extractor $f_{\theta}$ and a cross attention module $h_{\varphi}$ with learnable parameters $\theta$ and $\varphi$, which we will elaborate next. An illustration of our pipeline can be found in Fig.~\ref{fig:pipeline}.

\noindent\textbf{Embedding Extractor Module} computes per point intermediate embedding $\hat{\bPsi}_{(\cdot)}$, which is a non-linear mapping:

\begin{equation}
    f_{\theta} : \bV_{(\cdot)} \rightarrow \hat{\bPsi}_{(\cdot)}
\end{equation}

where $\cdot$ can either be shape $\cS$ or $\cT$ and $\hat{\bPsi}_{(\cdot)}$ will be further refined in the up-coming cross attention block.  

Note that many point cloud learning methods \cite{qi2016pointnet, qi2017pointnetpp, wang2019dgcnn, sharp2022diffusionnet} discussed in Sec.~\ref{subsec:rw_pcl_learning} can be employed here. 
However, careful design choice is required due to our special learning objective, namely the learned embedding must retain close to the LBO eigenbasis (cf. Sec.~\ref{subsection:method_loss}), indicating that the learned (intermediate) embedding must be smooth.
This relates to the fact that the smallest eigenfunctions (low frequency) of LBO vary smoothly on the manifold.

This naturally leads to the choice of As-Smooth-As-Possible (ASAP) DiffusionNet, a variant of DiffusionNet architecture proposed by Attaiki et al.~\cite{attaiki2023clover} as the default backbone of our embedding extractor.
It captures the local geometric information of different scales on the manifold by modelling a heat diffusion process with different timesteps and constrains the learned embedding to live in the space spanned by the LBO eigenbasis.
Both aspects encourage smoothness while retaining expressiveness of the learned embedding, which we found particularly suitable for our task.
Note that the realm of point cloud learning is still very active and yet our pipeline is flexible, that advances in the field can be directly incorporated by a drop-in replacement of the ASAP DiffusionNet.

\noindent\textbf{Cross Attention Block} refines the independently predicted intermediate embeddings $\hat\bPsi_{\cS}$, $\hat\bPsi_{\cT}$ by encouraging the communication between them. It follows the Transformer architecture \cite{vaswani2017attention} and learns a non-linear mapping:

\begin{equation}
    h_{\varphi} : \{\hat\bPsi_{\cS}, \hat\bPsi_{\cT}\} \rightarrow \{\bPsiS, \bPsiT\}
\end{equation}

The output $\bPsiS$ and $\bPsiT$ are directly used to form our unsupervised loss (cf. Sec.~\ref{subsection:method_loss}), which will be minimised and update the learnable network parameters through back-propagation.
Specifically, we construct a fully connected, bipartite graph that connects every point on the shape $\cS$ with every points on the shape $\cT$. Each node in the graph is assigned with the corresponding intermediate embedding learned by the embedding extractor. 
The core concept of cross attention is that it computes a similarity matrix between the key and query (transformed version of $\hat\bPsi_{\cS}$, $\hat\bPsi_{\cT}$), and makes use of it to weight the value (again a transformed version of $\hat\bPsi_{\cS}$ or $\hat\bPsi_{\cT}$) to produce the final output (please refer to \cite{vaswani2017attention, attaiki2021dpfm} for details).

The key in this process is that it enables the cross-talk of the intermediate embeddings $\hat\bPsi_{\cS}$ and $\hat\bPsi_{\cT}$, which is essential for a coupled and consistent shape embedding. 
%
This is akin to the idea of joint diagonalisation \cite{kovnatsky2013coupled} and image co-segmentation \cite{wang2013coseg}, where the information of the other object (shape/image) has to be made available in some way to achieve consistency.
As a result, our final embedding is aware of the other shape and hence coupled and consistent as shown in Fig.~\ref{fig:embedding} and Fig.~\ref{fig:supp_basis_m}.

\input{tables/table_near_isometric}

\subsection{Unsupervised Loss}
\label{subsection:method_loss}

Our unsupervised loss is inspired by the work of classical geometry processing \cite{kovnatsky2013coupled, eynard2015joint} and consists of three terms. 
Among them the off-diagonal loss and the orthogonal loss together encourage the learned embeddings to behave similarly as the classical LBO eigenbases, and the contrastive loss penalises their inconsistency.
We will introduce them one by one in the following.

\noindent\textbf{Off-diagonal Loss:} Similar as in Eq.~\eqref{eq:coupledDiag}, the learned embedding $\bPsi_{(\cdot)}$ should approximately diagonalise the respective Laplacian $\bL_{(\cdot)}$. 

\begin{equation}\label{eqn:off}
L_{\mathrm{off}} = \sum_{i \in \{\cS, \cT\}} \left\| \bPsi_i^{T} \bL_i \bPsi_i - \eval_i \right\|_F
\end{equation}

Note that $\eval_{(\cdot)}$ is a diagonal matrix of increasing eigenvalues of the respective LBO sitting on the diagonal. 
This term also encourages the learned embedding to be frequency-aligned, namely the smoother (lower frequency) an embedding is, the earlier it is positioned in the full set of embeddings.

\noindent\textbf{Orthogonal Loss:} The orthogonal constraint in Eq.~\eqref{eq:coupledDiag} is relaxed to a soft penalty in our training objective. 
It encourages the learned embedding to possess a basis structure and prevent undesired rank deficiency, hence maximising the embedding space spanned by the learned embeddings. 

\begin{equation}\label{eqn:ortho}
L_{\mathrm{o}} = \sum_{i \in \{\cS, \cT\}} \left\| \bPsi_i^\top \bM_i \bPsi_i - \bI \right\|_F
\end{equation}

In fact, the optimal embedding to minimise both the orthogonal loss and the off-diagonal loss is the individual LBO eigenbasis of shape $\cS$ and $\cT$, which is a special case of our formulation.
Moreover, we circumvent the intractable complexity of high-dimensional manifold optimisation in Eq.~\eqref{eq:coupledDiag} by leveraging a data-drive learning technique, which enables a direct prediction of per-point embedding without any subspace parameterisation required in \cite{kovnatsky2013coupled, eynard2015joint}. 

\noindent\textbf{Contrastive Loss:} This term couples the learned embeddings $\bPsiS$ and $\bPsiT$ and encourages their mutual consistency. 

\begin{equation}\label{eqn:contrastive}
L_{\mathrm{c}} = \left\| \bD_\cS^{T} \bM_\cS \bPsi_\cS - \bD_\cT^{T} \bM_\cT \bPsi_\cT \right\|_F
\end{equation}

Similar as in Eq.~\eqref{eq:coupledDiag}, the coupling is achieved by driving the Fourier coefficients of corresponding descriptor functions $\bD_{\cS}$ and $\bD_{\cT}$ to be as close as possible. 
Different to Eq.~\eqref{eq:coupledDiag}, we can learn highly accurate coupled embeddings from low-quality descriptor functions (f.e.~HKS), fully eliminating the need of ground truth correspondences required in \cite{kovnatsky2013coupled, eynard2015joint} (see Sec.~\ref{sec:exp}, Tab.~\ref{tab:complete_shape}). 
To our best knowledge, this enables, for the first time, the practical application of dense shape correspondence estimation based on coupled embeddings.

\noindent Finally, our full unsupervised loss is written as:
\begin{equation}\label{eqn:final}
L_{\mathrm{total}} = \mu_{\mathrm{off}}L_{\mathrm{off}} + \mu_{\mathrm{o}}L_{\mathrm{o}} + \mu_{\mathrm{c}}L_{\mathrm{c}}
\end{equation}
where $\mu_{\mathrm{off}}=1,\mu_{\mathrm{o}}=5e1$ and $\mu_{\mathrm{c}}=1e3$ are the corresponding weights. Please see supplementary for implementation details.

\subsection{Dense Correspondences}
\label{subsec:method_matching}
After the network is trained, we can directly obtain coupled embeddings $\bPsiS$, $\bPsiT$ from two input point clouds $\bVS$, $\bVT$ at inference time. 
Since both the coupled embeddings are predicted by the same network and live in the same embedding space, they are directly comparable. 
To retrieve dense pointwise correspondences, we employ the simple nearest neighbor search.

\begin{equation}
    \text{NN} : \{\bPsiS, \bPsiT\} \rightarrow \Pi_{\cS\cT}
\end{equation}

Namely for the $i$-th source point in shape $\cS$, we search for a target $j$-th point in $\cT$, whose $l_2$ distance to the source point is smallest in the embedding space and assign $\Pi_{\cS\cT}(i,j)=1$, indicating a match. 
Note that $\Pi_{\cS\cT}$ is a binary matrix, but not (always) a permutation matrix, since the correspondences are not guaranteed to be bijective.

%% file: tables/table_near_isometric.tex
\begin{table*}[t!]
    \scriptsize
    \centering
        \begin{tabular}{@{}lcccccc|ccccc@{}} 
        \toprule
        \multicolumn{1}{c}{\multirow{2}{*}{\textbf{Geo. error ($\times$100)}}} & \multicolumn{1}{c}{Train}  & \multicolumn{5}{c}{\textbf{FAUST}} &  \multicolumn{5}{c}{\textbf{SCAPE}}  \\
        & \multicolumn{1}{c}{Test} & \multicolumn{1}{c}{FAUST} & \multicolumn{1}{c}{SCAPE} & \multicolumn{1}{c}{SHREC19} & \multicolumn{1}{c}{TOPKIDS} & \multicolumn{1}{c}{DT4D-M} & \multicolumn{1}{c}{SCAPE} & \multicolumn{1}{c}{FAUST} & \multicolumn{1}{c}{SHREC19} & \multicolumn{1}{c}{TOPKIDS} & \multicolumn{1}{c}{DT4D-M} 

        \\ \midrule
        \multicolumn{1}{l}{NN Spectral Embedding} & & \multicolumn{1}{c}{67.1}  & \multicolumn{1}{c}{-} & \multicolumn{1}{c}{-} & \multicolumn{1}{c}{-} & \multicolumn{1}{c|}{-}  & \multicolumn{1}{c}{62.3} & \multicolumn{1}{c}{-} & \multicolumn{1}{c}{-} & \multicolumn{1}{c}{-} & \multicolumn{1}{c}{-} \\
        
        \multicolumn{1}{l}{HKS~\cite{sun2009concise}} & & \multicolumn{1}{c}{43.0}  & \multicolumn{1}{c}{-} & \multicolumn{1}{c}{-} & \multicolumn{1}{c}{-} & \multicolumn{1}{c|}{-}  & \multicolumn{1}{c}{40.5} & \multicolumn{1}{c}{-} & \multicolumn{1}{c}{-} & \multicolumn{1}{c}{-} & \multicolumn{1}{c}{-} \\ 

        \multicolumn{1}{l}{CQHB-HKS~\cite{kovnatsky2013coupled}} & & \multicolumn{1}{c}{37.2}  & \multicolumn{1}{c}{-} & \multicolumn{1}{c}{-} & \multicolumn{1}{c}{-} & \multicolumn{1}{c|}{-}  & \multicolumn{1}{c}{31.6} & \multicolumn{1}{c}{-} & \multicolumn{1}{c}{-} & \multicolumn{1}{c}{-} & \multicolumn{1}{c}{-} \\         
        
        \multicolumn{1}{l}{CQHB-GT~\cite{kovnatsky2013coupled}} & & \multicolumn{1}{c}{10.5}  & \multicolumn{1}{c}{-} & \multicolumn{1}{c}{-}  & \multicolumn{1}{c}{-} & \multicolumn{1}{c|}{-} & \multicolumn{1}{c}{10.8} & \multicolumn{1}{c}{-} & \multicolumn{1}{c}{-} & \multicolumn{1}{c}{-} & \multicolumn{1}{c}{-} \\

        \multicolumn{1}{l}{SyNoRiM(S)~\cite{huang2022multiway}} & & \multicolumn{1}{c}{7.9}  & \multicolumn{1}{c}{21.9} & \multicolumn{1}{c}{25.5} & \multicolumn{1}{c}{-} & \multicolumn{1}{c|}{-}  & \multicolumn{1}{c}{9.5} & \multicolumn{1}{c}{24.6} & \multicolumn{1}{c}{26.8} & \multicolumn{1}{c}{-} & \multicolumn{1}{c}{-} \\          

        \multicolumn{1}{l}{GeomFMaps(S)~\cite{donati20geomfmap}} & & \multicolumn{1}{c}{6.1}  & \multicolumn{1}{c}{11.2} & \multicolumn{1}{c}{10.8} & \multicolumn{1}{c}{26.2} & \multicolumn{1}{c|}{38.5}  & \multicolumn{1}{c}{7.7} & \multicolumn{1}{c}{9.0} & \multicolumn{1}{c}{12.4} & \multicolumn{1}{c}{21.7} & \multicolumn{1}{c}{28.6} \\ 
   
        \multicolumn{1}{l}{WSupFMNet(W)~\cite{sharma20weakly}} & & \multicolumn{1}{c}{6.0}  & \multicolumn{1}{c}{12.5} & \multicolumn{1}{c}{13.8}  & \multicolumn{1}{c}{28.9} & \multicolumn{1}{c|}{40.2} & \multicolumn{1}{c}{11.3} & \multicolumn{1}{c}{7.5} & \multicolumn{1}{c}{12.6} & \multicolumn{1}{c}{24.5} & \multicolumn{1}{c}{30.1} \\    
        
        \multicolumn{1}{l}{DiffFMaps(S)~\cite{marin2020lie}} & & \multicolumn{1}{c}{4.3}  & \multicolumn{1}{c}{18.7} & \multicolumn{1}{c}{14.6} & \multicolumn{1}{c}{20.5} & \multicolumn{1}{c|}{18.5}  & \multicolumn{1}{c}{14.4} & \multicolumn{1}{c}{10.8} & \multicolumn{1}{c}{14.2} & \multicolumn{1}{c}{18.0} & \multicolumn{1}{c}{15.9} \\        
        
        \multicolumn{1}{l}{NIE(W)~\cite{jiang2023nie}} & & \multicolumn{1}{c}{5.9}  & \multicolumn{1}{c}{16.7} & \multicolumn{1}{c}{15.1}  & \multicolumn{1}{c}{18.9} & \multicolumn{1}{c|}{13.3} & \multicolumn{1}{c}{11.6} & \multicolumn{1}{c}{8.6} & \multicolumn{1}{c}{13.2} & \multicolumn{1}{c}{16.2} & \multicolumn{1}{c}{12.1} \\       

        \multicolumn{1}{l}{NIE(W)~\cite{jiang2023nie}(with ASAP)} & & \multicolumn{1}{c}{5.6}  & \multicolumn{1}{c}{15.0} & \multicolumn{1}{c}{20.7}  & \multicolumn{1}{c}{19.7} & \multicolumn{1}{c|}{13.5} & \multicolumn{1}{c}{12.6} & \multicolumn{1}{c}{5.9} & \multicolumn{1}{c}{23.5} & \multicolumn{1}{c}{15.3} & \multicolumn{1}{c}{12.0} \\      

        \multicolumn{1}{l}{SSMSM(W)~\cite{cao2023multimodal}} & & \multicolumn{1}{c}{\textbf{2.4}}  & \multicolumn{1}{c}{\textbf{{6.8}}} & \multicolumn{1}{c}{\textbf{{9.0}}} & \multicolumn{1}{c}{\textcolor{blue}{14.2}} & \multicolumn{1}{c|}{\textbf{11.8}}  & \multicolumn{1}{c}{\textcolor{blue}{4.1}} & \multicolumn{1}{c}{4.1} & \multicolumn{1}{c}{\textbf{5.2}} & \multicolumn{1}{c}{\textcolor{blue}{12.3}} & \multicolumn{1}{c}{\textcolor{blue}{8.0}} \\

        \multicolumn{1}{l}{Ours w/o ASAP(W)} & & \multicolumn{1}{c}{3.9}  & \multicolumn{1}{c}{8.8} & \multicolumn{1}{c}{16.2} & \multicolumn{1}{c}{15.3} & \multicolumn{1}{c|}{14.0}  & \multicolumn{1}{c}{4.3} & \multicolumn{1}{c}{\textcolor{blue}{3.9}} & \multicolumn{1}{c}{13.1} & \multicolumn{1}{c}{14.6} & \multicolumn{1}{c}{10.9} \\      
        
        \multicolumn{1}{l}{Ours(W)} & & \multicolumn{1}{c}{\textcolor{blue}{3.7}}  & \multicolumn{1}{c}{\textcolor{blue}{8.7}} & \multicolumn{1}{c}{\textcolor{blue}{9.5}} & \multicolumn{1}{c}{\textbf{13.7}} & \multicolumn{1}{c|}{\textcolor{blue}{13.1}}  & \multicolumn{1}{c}{\textbf{3.2}} & \multicolumn{1}{c}{\textbf{3.7}} & \multicolumn{1}{c}{\textcolor{blue}{8.1}} & \multicolumn{1}{c}{\textbf{11.0}} & \multicolumn{1}{c}{\textbf{7.8}} \\

        \bottomrule
        \end{tabular} 
    \caption{Quantitative results on FAUST, SCAPE, SHREC19, TOPKIDS and DT4D-M. The \textbf{best} results are highlighted, and the \textcolor{blue}{second best} results are indicated in blue. All methods only take point clouds as input except the multimodal method SSMSM~\cite{cao2023multimodal}, which requires meshes. Ours outperforms all baselines, both classical and learning-based methods, and is comparable (if not superior) to SSMSM. Letters S,W in parentheses stand for supervised and weakly supervised respectively.
    }
    \label{tab:complete_shape}
\end{table*}

%% file: sec/5_experiment.tex
\section{Experiments}
\label{sec:exp}

We start this section by introducing the most relevant baselines in Sec.~\ref{subsec:exp_basline} before reporting experiment results on near-isometric and non-isometric matching in Sec.~\ref{subsec:exp_near-isometric}~\&~\ref{subsec:exp_non_isometric}. Then we study the generalisation ability (Sec.~\ref{subsec:exp_generalisation}) and robustness (Sec.~\ref{subsec:exp_robustness}) of our method due to their high practical relevance. Lastly as proof-of-concepts, we show that our learned embedding can be used for challenging partial shape matching (Sec.~\ref{subsec:exp_partial}) and segmentation (Sec.~\ref{subsec:exp_segmentation}).

\subsection{Baselines}
\label{subsec:exp_basline}
We compare our method with relevant baselines, including both axiomatic and learning-based methods.

\emph{CQHB} is an inspiring work by Kovnatsky et al.~\cite{kovnatsky2013coupled} using classical optimisation. We evaluate it in two different settings: namely with HKS or ground truth correspondences as indicator functions and report their results as \emph{CQHB-HKS} and \emph{CQHB-GT} respectively. 

\emph{DiffFMaps}~\cite{marin2020lie}, \emph{NIE}~\cite{jiang2023nie} and \emph{SyNoRiM}~\cite{huang2022multiway} are the SOTA methods to learn shape embeddings (or bases as in \emph{SyNoRim}) from point clouds. 
They are the most related to ours since all methods aim to learn a good shape embedding under challenging non-rigid deformations. Additionally, we report the results of the most competitive method \emph{NIE} using ASAP DiffusionNet as feature extractor.

\emph{GeomFMaps}~\cite{donati20geomfmap} and \emph{WSupFMNet}~\cite{sharma20weakly} are SOTA fmaps-based learning methods, they are relevant to ours since they only take point clouds as input and produce dense shape correspondences. 

Lastly, the SOTA multimodal learning method \emph{SSMSM}~\cite{cao2023multimodal} is also a fmaps-based shape matching method.
In addition to the input point cloud, it also requires the face information contained in meshes, where as ours only needs raw point clouds.
We showcase in the experiments that our proposed method, despite only having access to point clouds, performs on par (if not superior) with \emph{SSMSM}. 

\begin{figure}[t]
    \centering
    \includegraphics[width=0.45\textwidth]{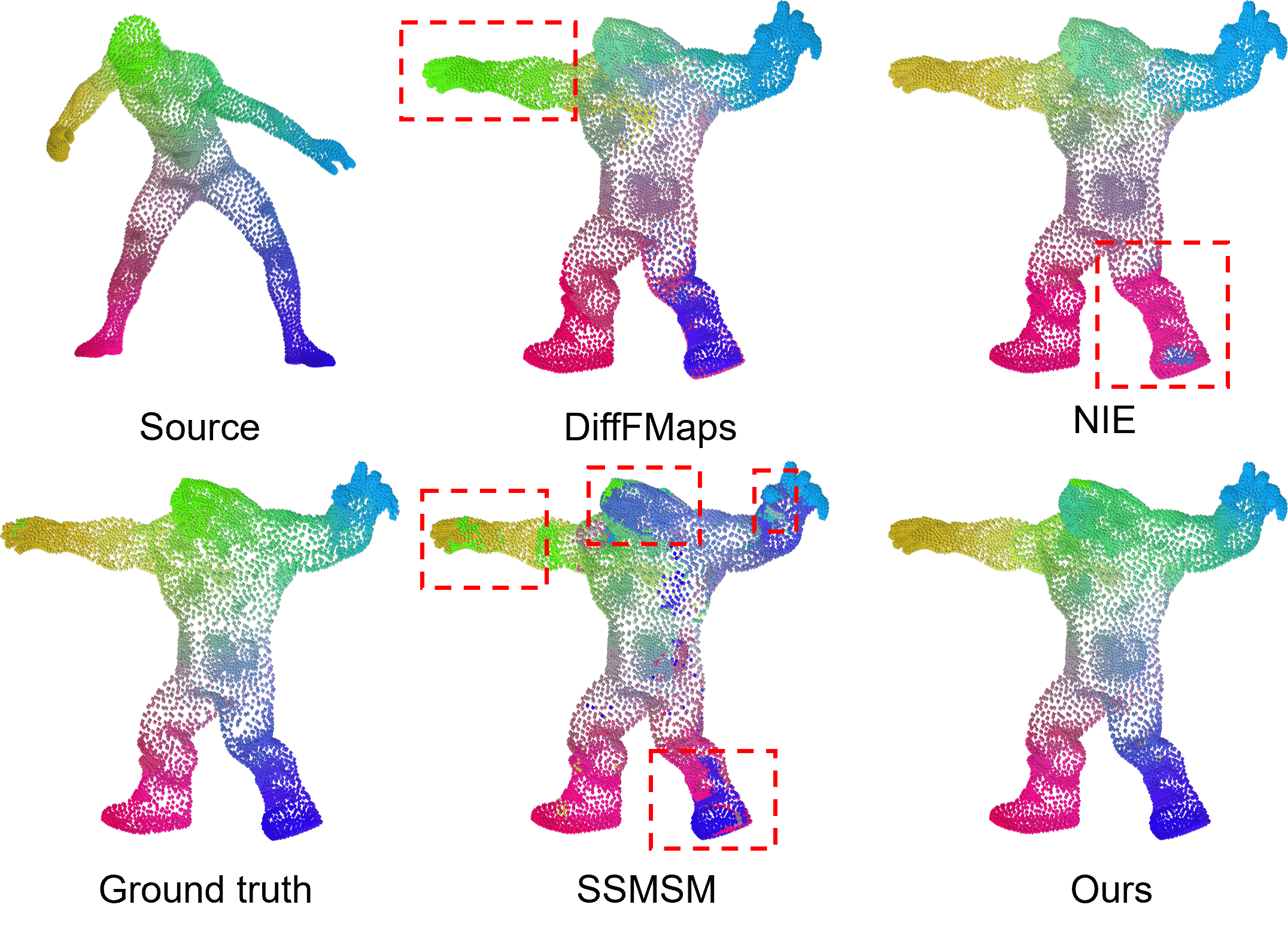}
    \caption{Qualitative result on DT4D-M. Ours produces the most accurate and smooth correspondences, despite highly non-isometric deformation (errors highlighted in red).}
    \label{fig:non_isometric}
    \vspace{-0.2cm}
\end{figure}

\subsection{Near-isometric Shape Matching}
\label{subsec:exp_near-isometric}

\noindent\textbf{Datasets}
We choose FAUST~\cite{bogo2014faust}, SCAPE~\cite{anguelov2005scape} and SHREC19~\cite{melzi2019shrec} as testbeds for the task of near-isometric shape matching, specifically the more recent remeshed version~\cite{ren2018continuous, donati2020deep} of them.
The FAUST dataset encompasses 100 human shapes, representing 10 individuals in 10 distinct poses. 
We split them as $80/20$ for train and test.
The SCAPE dataset comprises 71 shapes of a single person in different poses. 
We split them as $51/20$ for train and test.
The SHREC19 dataset includes 44 human shapes and is exclusively used as a test set.
Note that due to remeshing, the distribution of each point cloud is totally different, rendering the matching task more realistic and challenging.

\noindent\textbf{Results}
We train on FAUST and SCAPE respectively and evaluate on FAUST, SCAPE and SHREC19. 

As shown in Tab.~\ref{tab:complete_shape}, our proposed method can produce significantly more accurate correspondences than its classical counterpart CQHB, even under CQHB-GT in which dense ground truth correspondences are used. 
Moreover, we conduct a simple experiment to retrieve dense shape correspondences by nearest neighbor search directly using HKS (cf. Tab.~\ref{tab:complete_shape}). The quality of estimated correspondences is inferior.
However, our method can fully exploit the information available in the low-quality HKS descriptor and predict highly accurate correspondences.

Our method also outperforms SOTA learning methods, even the ones with ground truth supervision such as SyNoRiM and GeomFMaps. 
Remarkably, ours is capable to compete (it not superior) with the multimodal learning method SSMSM, which requires meshes as input. 
This highlights the importance of the careful design of our network architecture and our unsupervised loss inspired by the classical geometry processing technique. 
As an ablative study we disable the ASAP component hence employ the vanilla DiffusionNet as feature extractor and report its quantitative results in Tab.~\ref{tab:complete_shape} as Ours w/o ASAP. Note that the mean geodesic error deteriorates in all cases, underlining the importance of smoothness of learned embeddings.
Please refer to the supplementary for qualitative results and additional ablation experiments. 

\subsection{Non-isometric Shape Matching}
\label{subsec:exp_non_isometric}

\noindent\textbf{Datasets} We employ the recent non-isometric benchmark DT4D-M ~\cite{magnet2022smooth} as the testbed for this task. 
This dataset includes shapes from the large-scale animation dataset DT4D~\cite{li20214dcomplete} and consists of 293 humanoid shapes from 9 different classes. 
We split it as $198/95$ for train and test.
Following the train/test split proposed in~\cite{li2022attentivefmaps}, we conduct experiments with all 9 classes of humanoid shapes, which undergo significant non-isometric deformation (cf. Fig.~\ref{fig:non_isometric}).

\noindent\textbf{Results} We test on DT4D-M using our model trained on FAUST and SCAPE respectively. 
Note that this is a harder case than training and testing on the same dataset, since all methods are only trained with near-isometric shapes. 
However, our proposed method performs favorably than all baselines and achieves comparable result with mesh-dependent SSMSM method. 
We report the quantitative  and qualitative results in Tab.~\ref{tab:complete_shape} (column DT4D-M) and Fig.~\ref{fig:non_isometric} respectively.

\subsection{Generalisation}

\label{subsec:exp_generalisation}
\input{tables/table_large}

\label{subsec:generalisation}
\noindent\textbf{Datasets} To further study the generalisability of our proposed method, we employ the SURREAL dataset~\cite{varol2017learning}, which is a synthetic dataset of human shapes.
We train our model and baselines on a randomly sampled subset 
of the $230K$ synthetic shapes and test on FAUST, SCAPE, and SHREC19.

\begin{figure}[t!]
    \vspace{-0.2cm}
    \centering
    \includegraphics[width=0.45\textwidth]{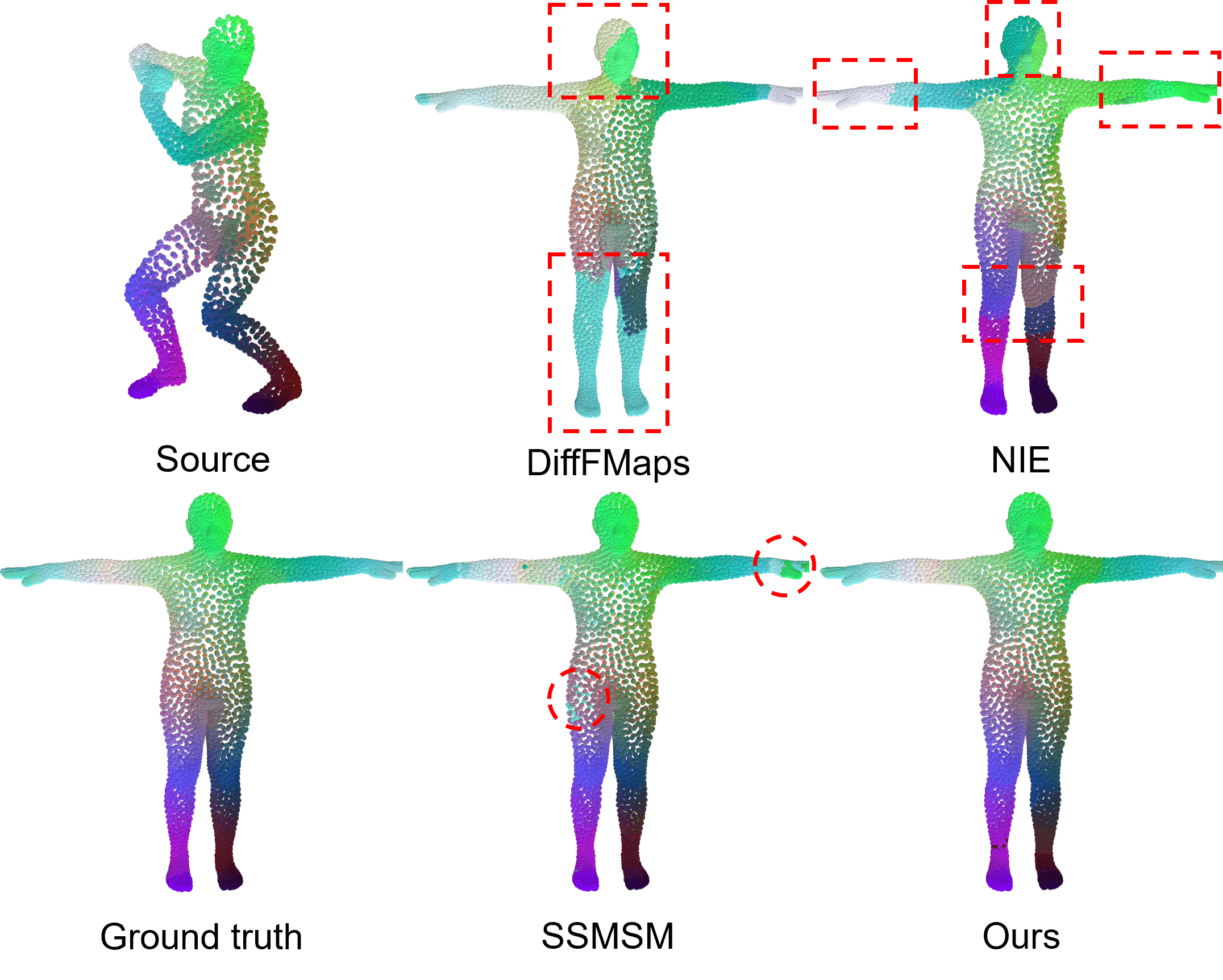}
    \caption{Generalisation from the training set SURREAL to the test set SHREC19. Our method generalises better compared to baselines (errors highlighted in red).}
    \label{fig:generalisation}
\end{figure}

\noindent\textbf{Results} Quantitative and qualitative results are reported in Tab.~\ref{tab:generalisation} and Fig.~\ref{fig:generalisation} respectively. Remarkably, ours outperforms all baselines including the multimodal mesh-dependent method SSMSM under this setting. A possible reason is that our learned embeddings are driven by the geometry-aware supervision, and are further coupled via both the network architecture (cross attention block) and the (constrastive) loss. This geometry-aware supervision and strong coupling foster the generalisation ability, leading to the superior performance of our proposed method.

\subsection{Robustness}
\label{subsec:exp_robustness}

We evaluate robustness from two perspectives: 
(1) random additive Gaussian noise to point clouds, 
(2) changes and inconsistency in shape topology.
Both scenarios are common in real-world raw point clouds, hence are highly relevant for the practicability of investigated methods.

\noindent\textbf{Additive Gaussian Noise} We make use of the trained model in SURREAL (Sec.~\ref{subsec:generalisation}) and test on \emph{noisy} point clouds from FAUST, SCAPE, SHREC19. Every point in the test point clouds is perturbed by a Gaussian with $\mu=0$ and $\sigma=0.01$ and within a range of $[-0.05, 0.05]$.
Quantitative results are shown in Tab.~\ref{tab:generalisation} (numbers in parentheses). Under this noisy setting, the quality of our correspondences retains the best among all competing methods. Compared to the noise-free case, we also have the least overall performance degradation. 
An illustration is shown in Fig.~\ref{fig:noisy} in the supplementary.

\noindent\textbf{Topology changes} We employ models pre-trained on FAUST and SCAPE respectively and test on the TOPKIDS dataset~\cite{lahner2016shrec}, which contains 26 shapes of kids with non-rigid deformation and topological changes for this task.

\begin{figure}[t]
    \centering
    \includegraphics[width=0.45\textwidth]{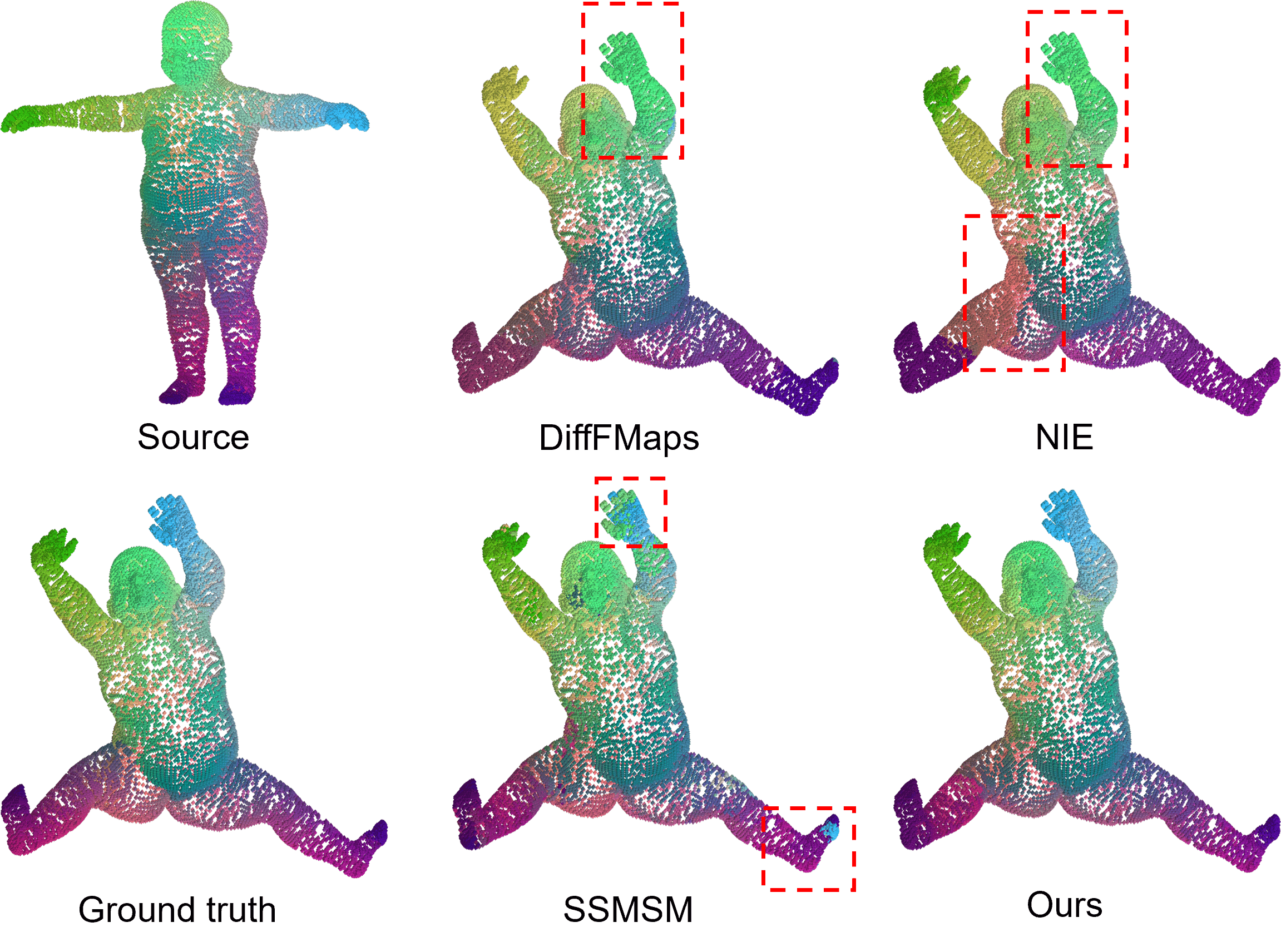}
    \caption{ Robustness against topological changes (the left shoulder and face of the kid are glued together). Ours is least sensitive to this noise among all competing methods (errors highlighted in red).
    }
    \label{fig:topology}
    \vspace{-0.3cm}
\end{figure}

Quantitative results are shown in Tab.~\ref{tab:complete_shape} (column TOPKIDS). Note that all investigated methods suffer from the challenging topological changes, however ours outperforms by achieving the lowest mean geodesic error. Qualitative illustration (Fig.~\ref{fig:topology} \& \ref{fig:supp_topkids}) also shows that our predicted correspondences are the closest to the ground truth. 

\subsection{Partial Shape Matching}
\label{subsec:exp_partial}
As a proof-of-concept, we show that our proposed method can be applied to the challenging partial shape matching. For this we train our network on SHREC16 Partiality~\cite{cosmo2016shrec}. During training we take a full and partial pair and employ an extended loss  (see supplementary for details). Once the network is trained, it can be used to directly match two partial shapes by proximity search in the embedding space. Examples can be found in Fig.~\ref{fig:teaser} and in the supplementary.

\subsection{Shape Segmentation}
\label{subsec:exp_segmentation}
As a proof-of-concept, we show that our learned embedding can also be employed for shape segmentation tasks. Specifically we conduct a k-mean clustering on the learned embedding of each shape. An example is shown in Fig.~\ref{fig:teaser} and the segmentation is meaningful and even consistent across different shapes, despite independently segmented.

%% file: tables/table_large.tex
\begin{table}[t!]
    \scriptsize
    \centering
        \begin{tabular}{@{}lcccc@{}}
        \toprule
        \multicolumn{1}{c}{\multirow{2}{*}{\textbf{Geo. error ($\times$100)}}}  & \multicolumn{1}{c}{Train}     & \multicolumn{3}{c}{\textbf{SURREAL}}  \\        
        & \multicolumn{1}{c}{Test}   & \multicolumn{1}{c}{FAUST}   & \multicolumn{1}{c}{SCAPE}  & \multicolumn{1}{c}{SHREC19} \\
        \midrule
        \multicolumn{1}{l}{GeomFMaps(S)~\cite{donati2020deep}} & & \multicolumn{1}{c}{10.4 (-)} & \multicolumn{1}{c}{8.7 (-)} & \multicolumn{1}{c}{14.1 (-)}   \\
        \multicolumn{1}{l}{WSupFMNet(W)~\cite{sharma20weakly}} & & \multicolumn{1}{c}{16.0 (-)} & \multicolumn{1}{c}{14.7 (-)}  & \multicolumn{1}{c}{27.8 (-)}  \\
        \multicolumn{1}{l}{DiffFMaps(S)~\cite{marin2020lie}}& & \multicolumn{1}{c}{7.8 (22.8)} & \multicolumn{1}{c}{18.9 (26.9)} & \multicolumn{1}{c}{27.8 (34.2)}   \\        
        \multicolumn{1}{l}{NIE(W)~\cite{jiang2023nie}} & & \multicolumn{1}{c}{6.9 (11.3)}  & \multicolumn{1}{c}{11.0 (17.2)} & \multicolumn{1}{c}{11.3 (18.2)}  \\  
        \multicolumn{1}{l}{NIE(W)~\cite{jiang2023nie}(with ASAP)} & & \multicolumn{1}{c}{5.5 (8.8)}  & \multicolumn{1}{c}{9.5 (13.6)} & \multicolumn{1}{c}{11.0 (16.3)}  \\  
        \multicolumn{1}{l}{SSMSM(W)~\cite{cao2023multimodal}} & & \multicolumn{1}{c}{3.5 (6.8)}  & \multicolumn{1}{c}{3.8 (6.4)} & \multicolumn{1}{c}{6.6 (9.8)} \\ 
        
        \multicolumn{1}{l}{Ours(W)(w/o ASAP)} & & \multicolumn{1}{c}{\textbf{3.3 (5.1)}}  & \multicolumn{1}{c}{4.3 (5.8)} & \multicolumn{1}{c}{8.8 (13.2)} \\      
        \multicolumn{1}{l}{Ours(W)} & & \multicolumn{1}{c}{3.4 (5.2)}  & \multicolumn{1}{c}{\textbf{3.3 (5.2)}} & \multicolumn{1}{c}{\textbf{4.6 (9.4)}} \\         
        \bottomrule
        \end{tabular} 
    \caption{Generalisation ability. The \textbf{best} results in each column are highlighted. Our method outperforms all learning based baselines. Letters S,W in parentheses stand for supervised and weakly supervised respectively.}
    \label{tab:generalisation}
\end{table}

%% file: sec/6_conclusion.tex
\section{Limitations, Future Work and Conclusion}

In this paper, we proposed an unsupervised method to learn high-quality, well-generalised embeddings directly from raw point clouds. The embedding is aware of the underlying shape geometry and robust to various shape artefacts and non-rigid (both isometric and non-isometric) deformations and can be used to obtain dense correspondences via a simple proximity search in the canonical embedding space. 
Extensive experiments showcase that our proposed method achieves superior results in a number of non-rigid matching benchmarks and is promising in other shape analysis challenges, such as partial shape matching and segmentation, hence setting the new state-of-the-art.
Our method also has limitations. First, it requires shapes to be pre-aligned. An interesting direction is to incorporate the advancement in SO(3)/SE(3) invariant architecture~\cite{deng2021vn} to eliminate the necessity of pre-alignment.
Second, it is interesting to explore the possibility for a fully descriptor-free approach.
Lastly, an extension of our method to shape collections would be a promising avenue for future research.

\section{Acknowledgment}
The project is supported by the ERC Advanced Grant SIMULACRON and Munich Center for Machine Learning.

%% file: sec/X_suppl.tex
\clearpage
\setcounter{page}{1}

\maketitlesupplementary

In this supplementary material, we aim to provide additional implementation details, ablation studies and more experimental results. The document is organised as follows:
first, we provide details about Laplace-Beltrami Operator in Sec.~\ref{sec:supp_lbo} and implementation details including necessary pre-processing steps in Sec.~\ref{sec:supp_imp}. 
In Sec.~\ref{sec:supp_ablation} we ablate our our architecture and loss design and show an example how a challenging case is solved step-by-step by our careful design (Sec.~\ref{subsec:supp_ablation_arch_loss}).
We further ablate the performance of our proposed method at different spectral resolutions, namely the dimension of our predicted embeddings  (Sec.~\ref{subsec:supp_ablation_k}). Moreover, we analyse the performance of our method's execution time with inputs of different sizes (Sec.~\ref{subsec:supp_ablation_size}).
Then we elaborate the extension of our method to the partial setting in Sec.~\ref{sec:supp_partial}.
Finally, we present more qualitative results on various datasets and failure cases in Sec.~\ref{sec:supp_vis}.

\section{Laplace-Beltrami Operator}
\label{sec:supp_lbo}
For a given function $u$ defined on a Riemannian manifold $\mathcal{M}$, the Laplace-Beltrami Operator $\Delta u$ measures how the function deviates from its average value within each local neighborhood, taking into account the geometry of $\mathcal{M}$. 
This property leads to its widespread application in computational geometry and computer graphics, especially when dealing with curved surfaces or manifolds.
In practical applications, the discrete LBO approximates the continuous operator, enabling its use on graphs or discrete meshes.

A prevalent variant of the discrete LBO is the cotangent Laplacian, which is widely used due to its ability to approximate the LBO on discrete surfaces. It is defined based on the cotangent values of the angles around each vertex. The discrete formulation of the LBO at a single vertex $i$ is represented as:
\begin{align}
\label{eq:discrete_single}
(\Delta u)_{i} &\approx \frac{1}{2\cA_i} \sum_{j \in N(i)} (\cot \alpha_{ij} + \cot \beta_{ij})(u_{i} - u_{j})
\end{align}
Here, $\cA_i$ denotes the area associated with vertex $i$, typically computed as one-third of the areas of the adjacent faces, and $\alpha_{ij}$ and $\beta_{ij}$ are the angles opposite to the edge connecting vertices $i$ and $j$. Based on the definition of the LBO: $\Delta = \bM^{-1} \bL$, the mass matrix $\bM$ and the stiffness matrix $\bL$ are defined as:
\begin{align}
\bM &= \textbf{diag}(\cA_1, \cA_2, \ldots, \cA_{n_{\cS}}) \\
\bL_{ij} &=
\begin{cases}
\omega_{ij} = -\frac{1}{2} (\cot \alpha_{ij} + \cot \beta_{ij}) & j \in \cN(i) \\
- \sum\limits_{j \in \cN(i)} \omega_{ij} & j = i \\
0 & \text{otherwise}
\end{cases}
\end{align}

Finally, the eigenvalues $\eval$ and eigenvectors $\bPhi$ of the LBO $\Delta$ can be calculated using the following formula:
\begin{align}
\bL \bPhi &= \bM \eval \bPhi
\end{align}

\section{Implementation Details}
\label{sec:supp_imp}
Our network is implemented in PyTorch~\cite{paszke2019pytorch}. The embedding extractor is based on the ASAP DiffusionNet with the default configuration published in \cite{attaiki2023clover} and the cross attention block is based on the implementation in \cite{attaiki2021dpfm}.
We employ HKS \cite{sun2009concise} as descriptor functions $\bD_{(\cdot)}$ in our pipeline, where we set its feature dimension $d=512$. 
The dimension of the predicted (both intermediate and final) embeddings $\bPsi_{(\cdot)}$ is set to be $k=50$. 
Note that the same configuration is used across all our experiments to ensure a non-biased comparison. 
For more details, please refer to the supplementary material.
The hyper-parameters in Eq.~\eqref{eqn:final} are chosen as follows: $\mu_{\mathrm{off}} = 1, \mu_{\mathrm{o}} = 5e1$ and $\mu_{\mathrm{c}} = 1e3$. 
Our network is trained using the Adam optimiser with a learning rate of $1e\text{-}3$ with a batch size of $1$, except for SURREAL the batch size is set to $4$.

As pre-processing, we approximately pre-align (using procrustes analysis or manually in blender) and normalise all shapes to the unit ball in each dataset, compute their point cloud LBOs \cite{Sharp2020LNT} and eigen-decomposite them to obtain the eigenvalues and eigenvectors, based on which the HKS is pre-computed as well.

\section{Ablation Study}
\label{sec:supp_ablation}
\subsection{Architecture and Loss}
\label{subsec:supp_ablation_arch_loss}
\input{tables/table_ablation}

In this section, we first explain the difference between DiffusionNet and our ASAP variant.Then we evaluate our proposed loss terms $L_{\mathrm{off}}$, $L_{\mathrm{o}}$ and $L_{\mathrm{c}}$. We discard $L_{\mathrm{off}}$ and $L_{\mathrm{o}}$ together, and then $L_{\mathrm{o}}, L_{\mathrm{c}}$ respectively, since $L_{\mathrm{off}}$ and $L_{\mathrm{o}}$ collectively encourages the desired basis structure. Next, we progressively disable the ASAP operation (by reverting to the original implementation of DiffusionNet) and the cross attention block to demonstrate the effectiveness of our proposed architecture design. Additionally, we evaluate the scenario of removing all cross-communication components: $L_{\mathrm{c}}$ and the cross attention block. 
All experiments follow the same configuration as in Sec.~\ref{subsec:exp_near-isometric} and are evaluated on FAUST and SCAPE.

Compared to the original DiffusionNet, the ASAP version achieves smoother features by projecting the embedding $\hat{\bPsi}^{'(i)}$ in layer $i$ onto the Laplacian basis during the diffusion process in each layer of the diffusion block, and then projecting it back to the original space. This operation encourages the output embedding $\hat{\bPsi}^{(i)}$ of each layer to approximate the properties of a smooth function, meaning that the embedding space can be better described using only low-frequency information. Specifically, this can be expressed as: $\hat{\bPsi}^{(i)} = \bPhi \bPhi^{\dagger} \hat{\bPsi}^{'(i)}$.

\begin{figure}[hpbt]
    \centering
    \includegraphics[width=0.43\textwidth]{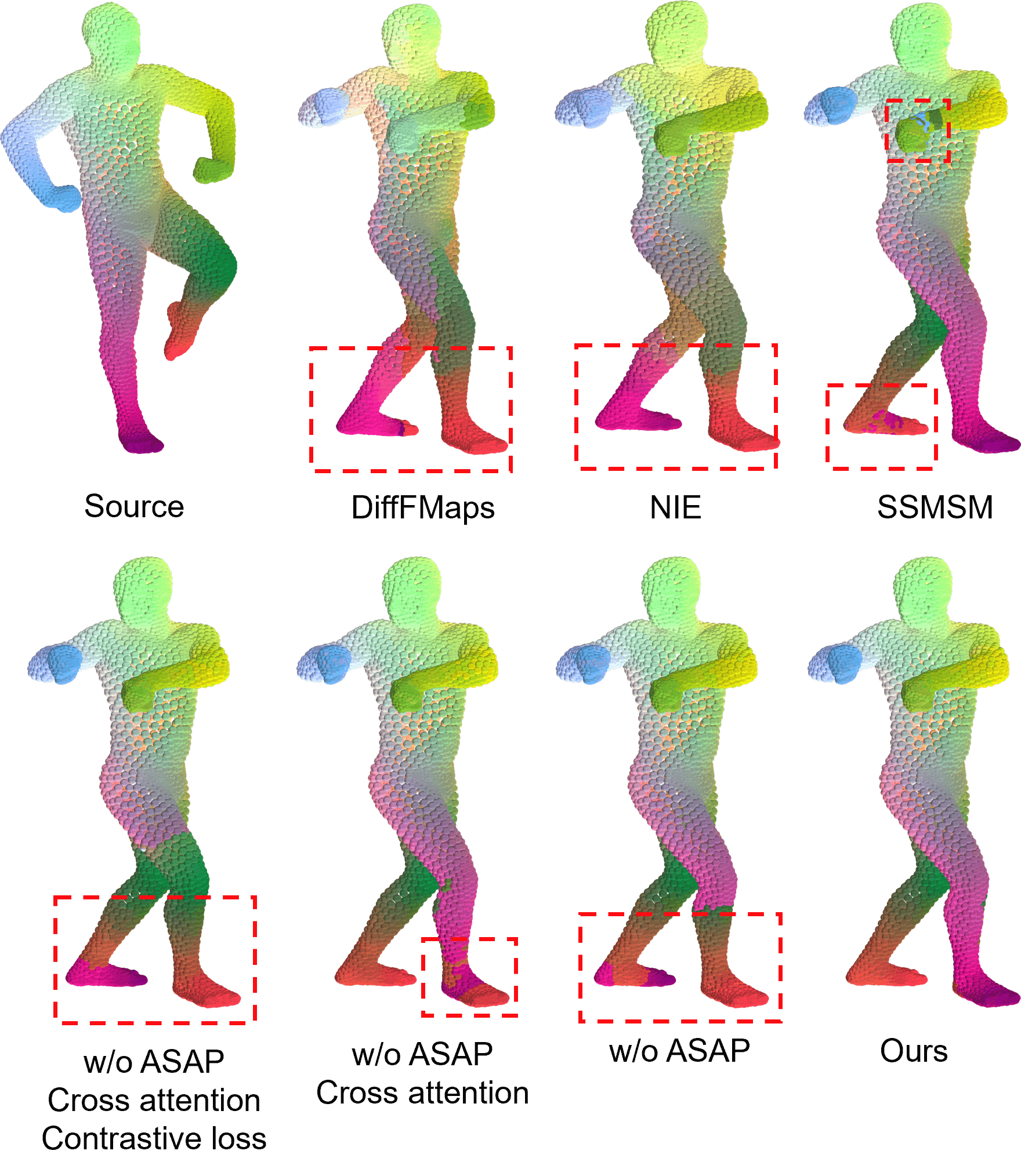}
    \caption{Visualisation of a challenging pair with crossed legs. We show our full design can successfully handle this challenge while all baseline methods fails (errors are highlighted in red).}
    \label{fig:ablation}
\end{figure}

\noindent\textbf{Results}
The quantitative results are reported in Tab.~\ref{tab:ablation}. Each loss term contributes to the accuracy of predicted correspondences. The $L_{\mathrm{c}}$ and the cross attention block facilitate communication cross shapes. The additional smoothness in the embedding induced by ASAP operation is also helpful. Fig.~\ref{fig:ablation} illustrates that our design addresses the challenging case of crossed legs step-by-step.

\subsection{Dimension of Learned Embeddings}
\label{subsec:supp_ablation_k}

\begin{figure}[t!]
    \centering
    \includegraphics[width=0.45\textwidth]{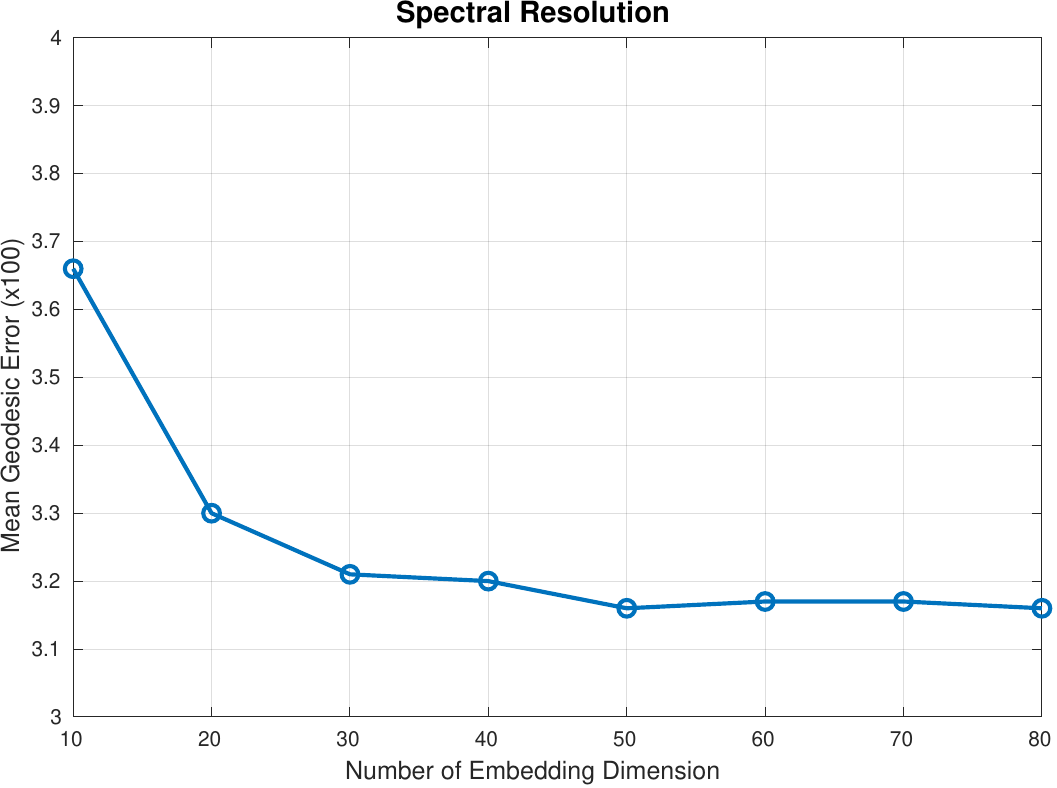}
    \caption{Illustration of mean geodesic error under different spectral resolutions. Our method is robust for different choice of spectral resolution.}
    \label{fig:spectral}
\end{figure}

We analyze the sensitivity of the dimension $k$ of our learned embeddings. In this experiment, we change $k$ gradually from $10$ to $80$ and train in total $8$ different networks on the SCAPE dataset.
As shown in Fig.~\ref{fig:spectral}, with the increase of $k$, the mean geodesic error of our predicted dense correspondences decrease rapidly initially, then stagnates stably at a low level. This is expected since we design our off-diagonal loss such that the learned embeddings are ``frequency-aligned", enabling a relatively faithful representation of the shape already with a handful of embeddings.

\subsection{Runtime Complexity}
\label{subsec:supp_ablation_size}

\begin{figure}[t!]
    \centering
    \includegraphics[width=0.45\textwidth]{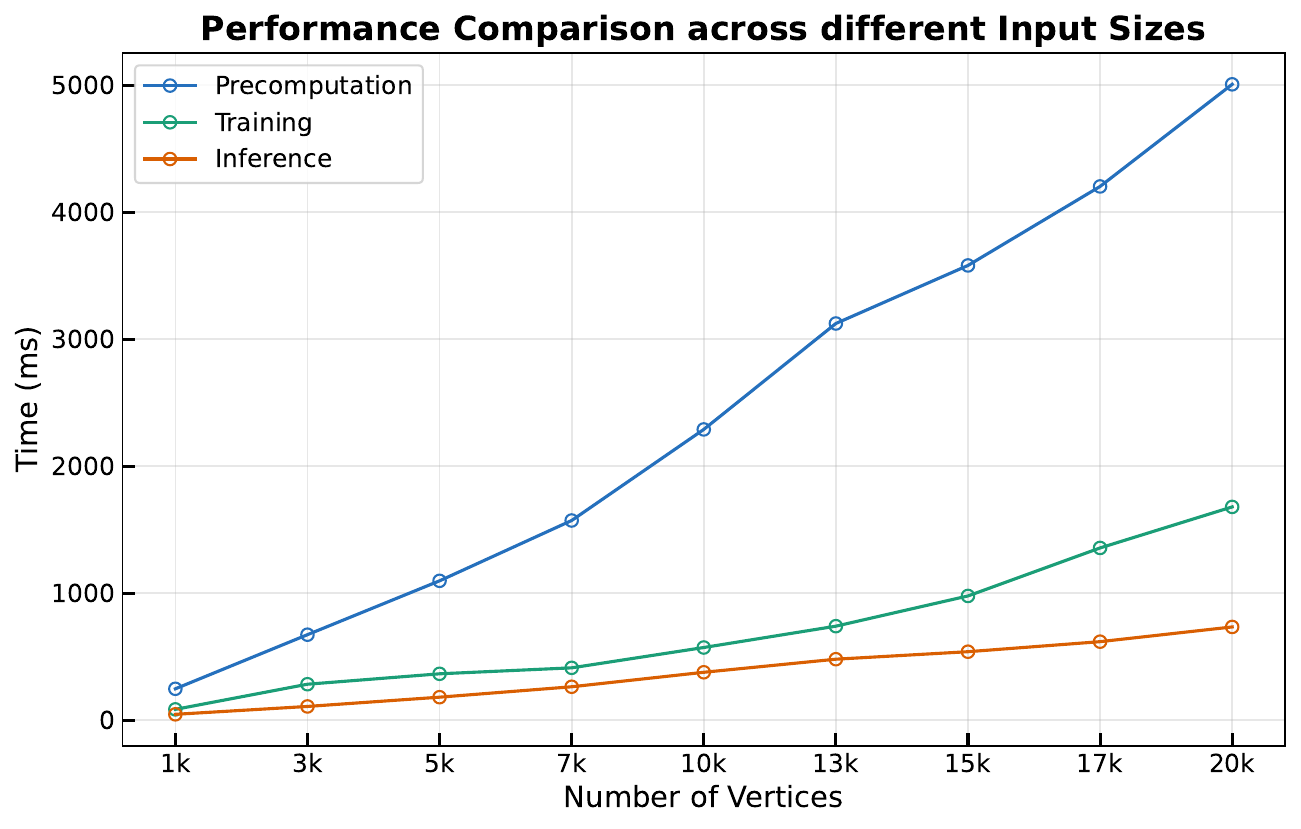}
    \caption{Our runtime vs. input point clouds size, ranging from 1k to 20k vertices. The runtime is measured per pair of shapes per forward pass, as our network processes two shapes simultaneously.}
    \label{fig:performance_comparison}
\end{figure}

In this section, we analyse the execution time and scalability of our approach with respect to input size. Specifically, we divide the total execution time into three stages: Precomputation, Training, and Inference. We conduct this analysis using 20 shapes sampled from the FAUST dataset, measuring the average run-time for each stage on an NVIDIA A800 GPU and an Intel Xeon Gold 6348 CPU @ 2.60GHz.

\begin{figure}[hpbt]
    \centering
    \includegraphics[width=0.3\textwidth]{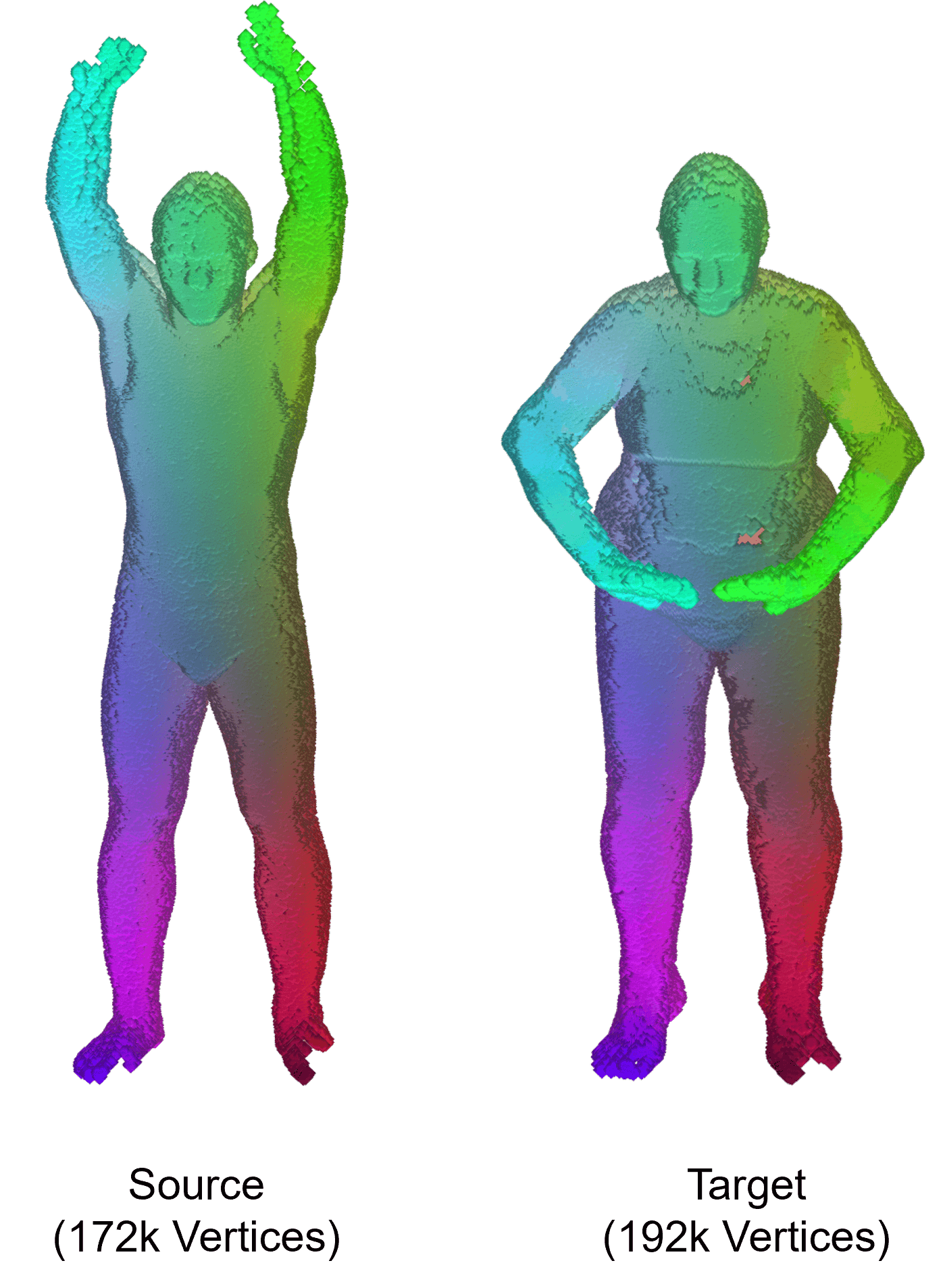}
    \caption{Matching results on MPI-FAUST~\cite{FAUST} raw scan data. We downsample the original point cloud to 50k for better visualization. Our method can correctly handle point clouds of large sizes.}
    \label{fig:supp_size}
\end{figure}

Fig.~\ref{fig:performance_comparison} presents the run-time across different input sizes and stages. Although preprocessing involves some computational overhead, our method remains efficient and scalable, providing robust performance even for large and complex inputs. Furthermore, Fig.~\ref{fig:supp_size} demonstrates our approach's capability to effectively match large-scale point clouds containing approximately 180k vertices.

\section{Partial Shape Matching}
\label{sec:supp_partial}

As a proof-of-concept, we extend our method to the challenging task to match partial shapes. We take a full-partial pair (from the same shape category in SHREC16~\cite{cosmo2016shrec}) during the training, and once the network is trained, we can directly match two partial shapes at inference time.

For this we introduce a new off-diagonal term for partial shapes analogous to the original one discussed in the main paper (cf. Eq.~\eqref{eqn:off}). 

\begin{equation}\label{eqn:off_partial}
L_{\mathrm{off\_partial}}= \left\| \bPsi^{T} \bL \bPsi - \textbf{diag}(\bPsi^{T} \bL \bPsi) \right\|_F
\end{equation}

This adapted loss term only asks for the off-diagonal term of $\bPsi^{T} \bL \bPsi$ to be as small as possible without any preference of frequency-alignment. This is reasonable because  Eq.~\eqref{eqn:off} will still be applied to the full shape and the frequency should be dictated by the full shape alone due to missing eigenvalues and eigenvectors under shape partiality~\cite{rodola2017partial}. Further we disable the orthogonal loss on the predicted embeddings of partial shapes. The reason behind is the embedding should be coupled to the one from the full shape, which contains frequencies that are absent in the partial shape. Fundamentally, it relates to deleting rows (or columns) of a Stiefel matrix will break the orthogonality, leading to an matrix which is not Stiefel anymore. The final loss reads as follows:


\begin{align}\label{eqn:final_partial_full}
L_{\mathrm{final}} = &\ \mu_{\mathrm{off\_partial}}L_{\mathrm{off\_partial}} 
+ \mu_{\mathrm{off\_full}} L_{\mathrm{off\_full}} \nonumber \\
&\ + \mu_{\mathrm{o\_full}}L_{\mathrm{o\_full}} 
+ \mu_{\mathrm{c}}L_{\mathrm{c}}.
\end{align}

We employ deep features extracted from SSMSM \cite{cao2023multimodal} (instead of HKS~\cite{sun2009concise}). 
This choice was made because heat diffusion behaves differently under different partiality, resulting in even lower SNR of the final HKS, making it much harder to learn anything useful from it.
%
Since our method is weakly supervised, we can exploit a test-time adaptation as in~\cite{cao2023unsupervised}. 
The hyperparameters are set as follows: 
$\mu_{\mathrm{off\_partial}} = \mu_{\mathrm{off\_full}} = 1, \mu_{\mathrm{o\_full}} = 5e3 $ and $\mu_{\mathrm{c}} = 5e3$.


\begin{figure}[hbt!]
    \centering
    \includegraphics[width=0.45\textwidth]{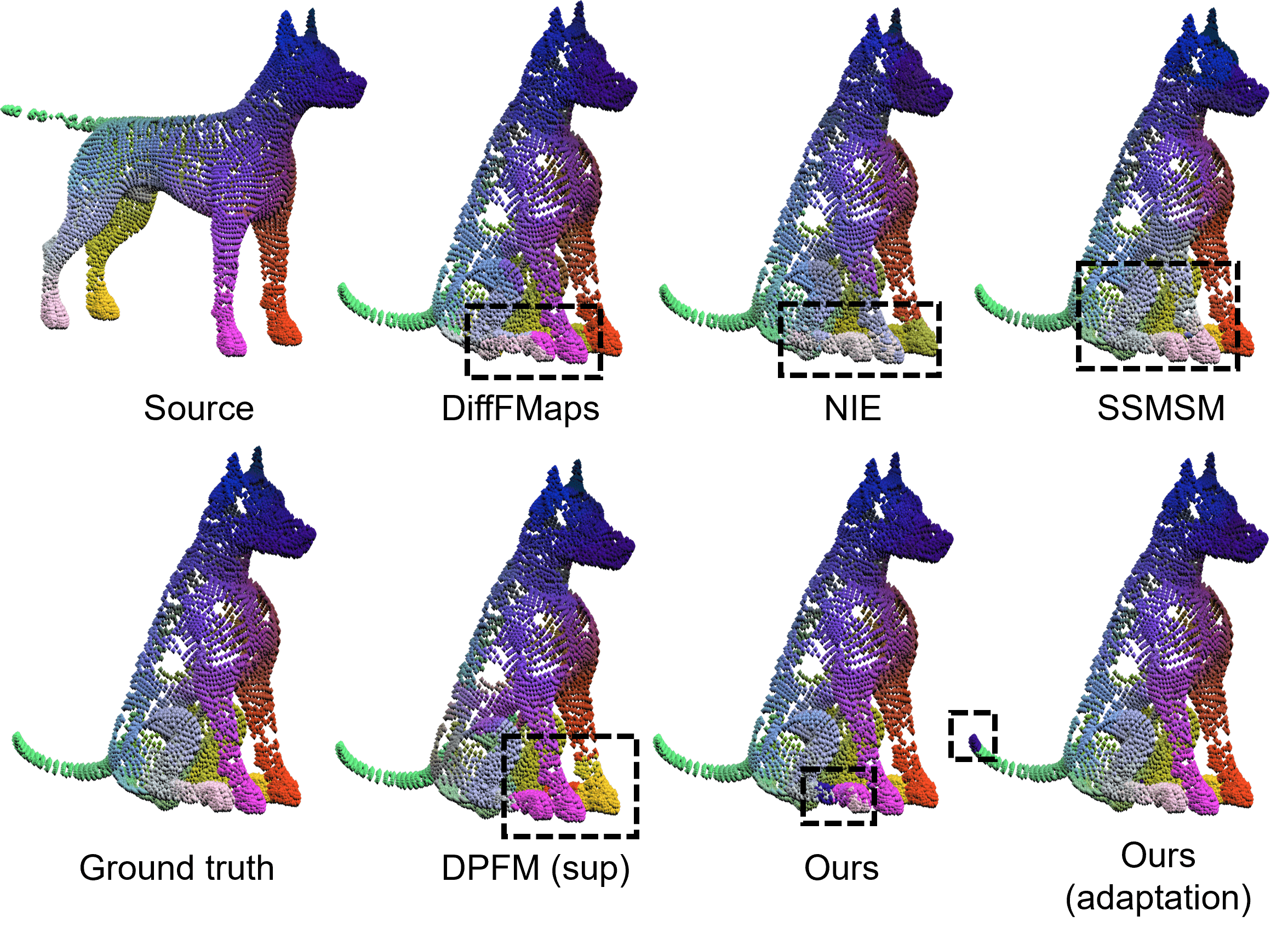}
    \caption{Qualitative result on SHREC16 HOLES. Ours performs reasonably well under this challenging setting.}
    \label{fig:partial}
\end{figure}

We conduct experiments using the SHREC16 partiality dataset~\cite{cosmo2016shrec} and our proposed method show promising results especially in the HOLES sub-dataset (cf. Fig.~\ref{fig:partial}). However we leave a comprehensive study as future work as partiality is challenging and deserves a thorough discuss itself.

\section{More Qualitative Results}
\label{sec:supp_vis}
In this section, we present additional qualitative results including failure cases. See figure captions for explanation.

\begin{figure*}[t!]
    \centering
    \includegraphics[width=1\textwidth]{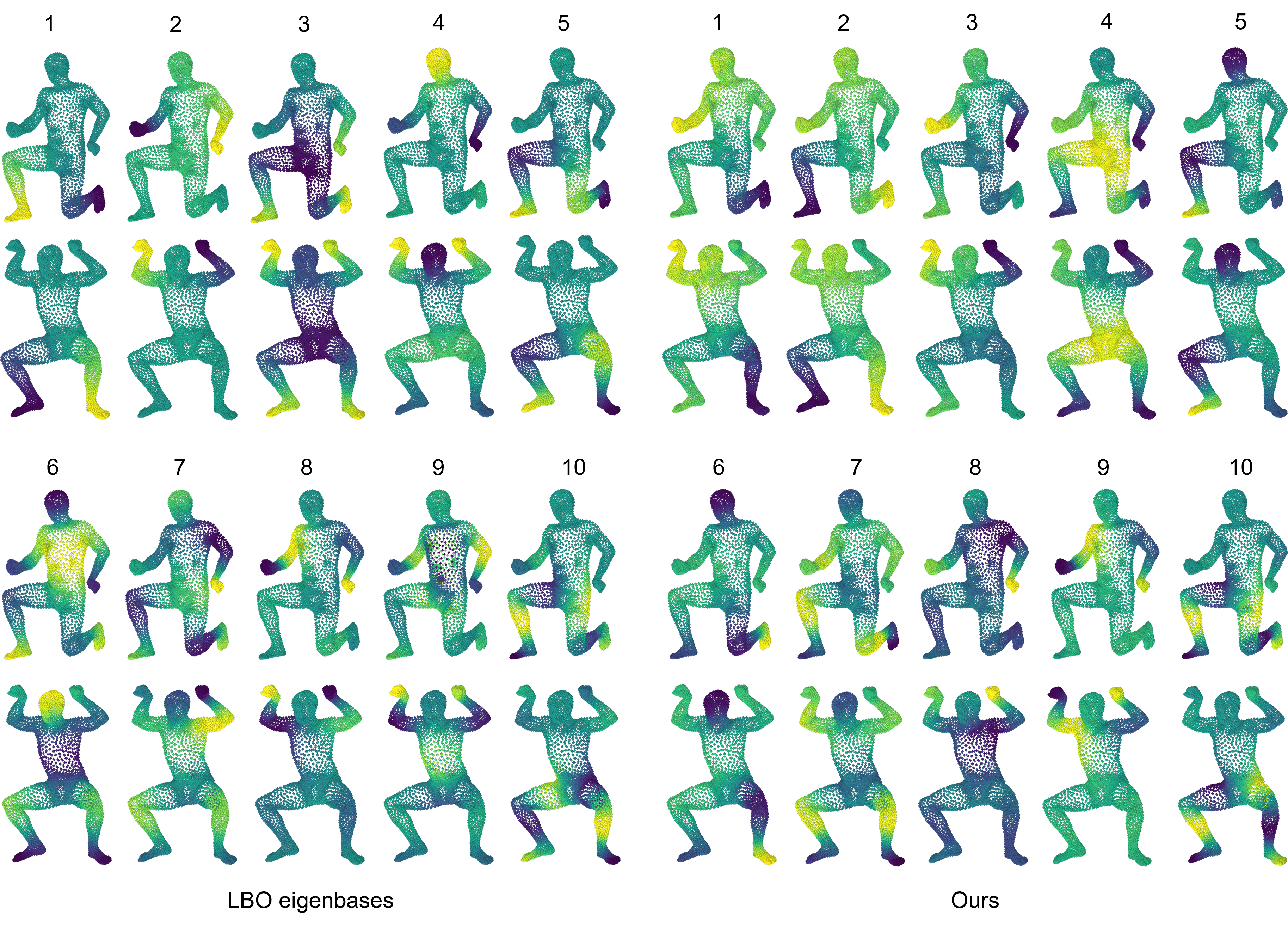}
    \caption{Visual comparison of our learned coupled embeddings vs. the LBO eigenbases. The first ten are shown. It can be seen that our learned embeddings are consistent while the LBO eigenbases suffer from sign flips and ambiguity in space corresponding to repeated eigenvalues.}
    
    \label{fig:supp_basis_m}
\end{figure*}

\begin{figure*}[b!]
    \centering
    \includegraphics[width=0.65\textwidth]{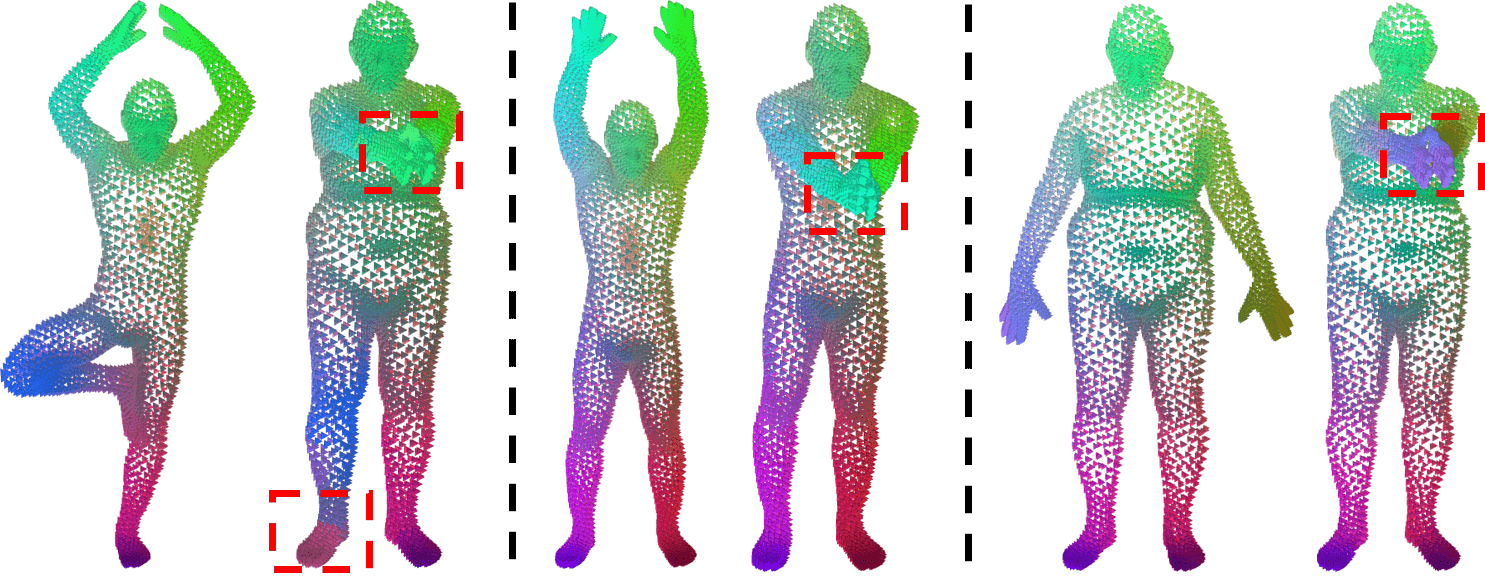}
    \caption{Failure cases on FAUST. All three failure examples relate to the touching hands, where the points of two hands are locally mixed and hard to separate. Note that this is the most challenging case for all point cloud methods in FAUST.}
    \label{fig:supp_faust_failure}
\end{figure*}

\newpage

\begin{figure*}[t!]
    \centering
    \includegraphics[width=0.9\textwidth]{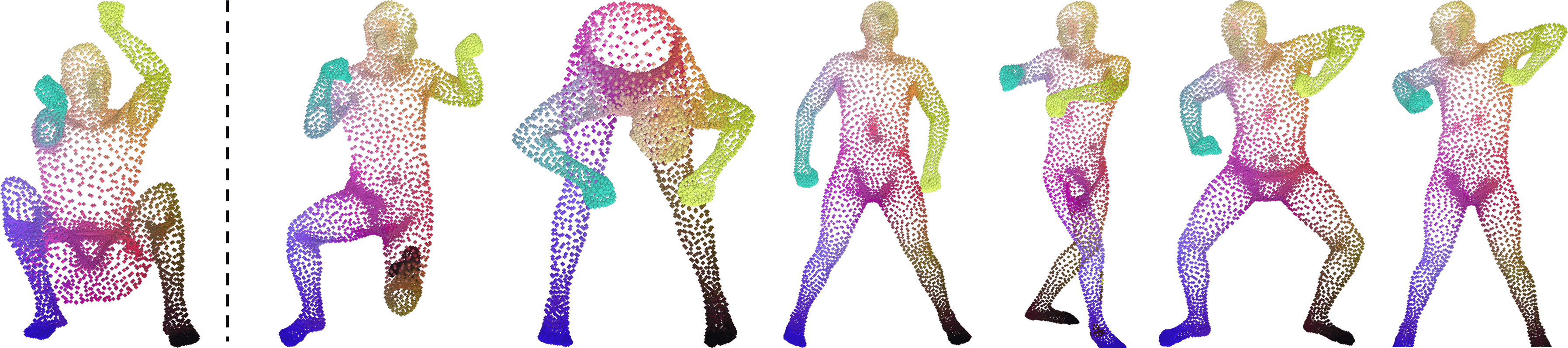}
    \caption{Qualitative results on SCAPE. Leftmost is the reference shape. Accurate correspondences are consistently obtained by our proposed method.}
    \label{fig:supp_scape}
\end{figure*}


\begin{figure*}[h]
    \centering
    \includegraphics[width=1\textwidth]{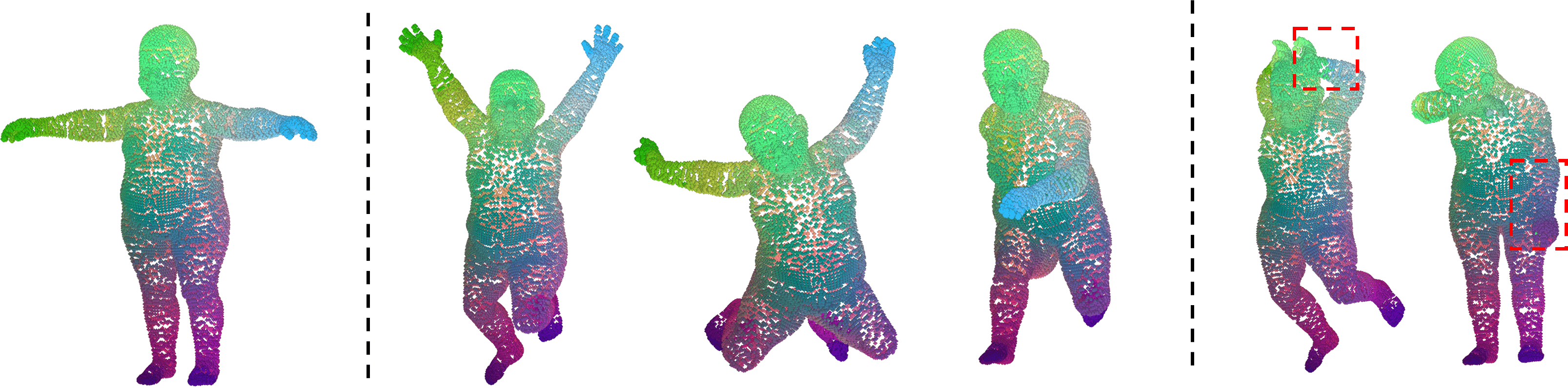}
    \caption{Qualitative results on TOPKIDS. Leftmost is the reference shape. Although our method can handle topological noise (\emph{middle}) , it still suffers from significant topological changes (\emph{right}).
    }
    \label{fig:supp_topkids}
\end{figure*}

\begin{figure*}[b!]
    \centering
    \includegraphics[width=0.9\textwidth]{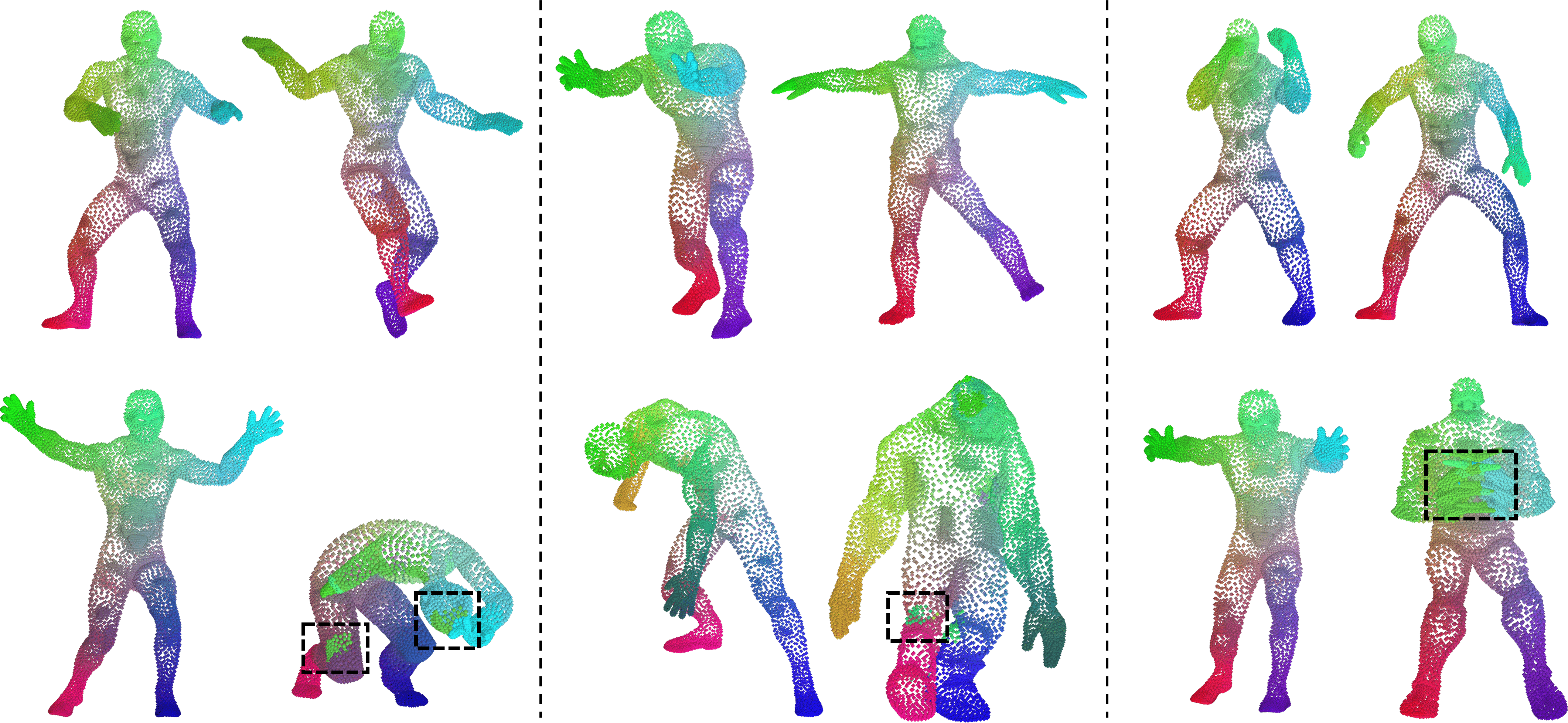}
    \caption{Qualitative results on DT4D-M. More qualitative non-isometric matching results (\emph{top}) . Failure cases mainly due to challenging topological noise (\emph{bottom}) .}
    \label{fig:supp_dt4d}
\end{figure*}



\begin{figure*}[t!]
    \includegraphics[width=1\textwidth]{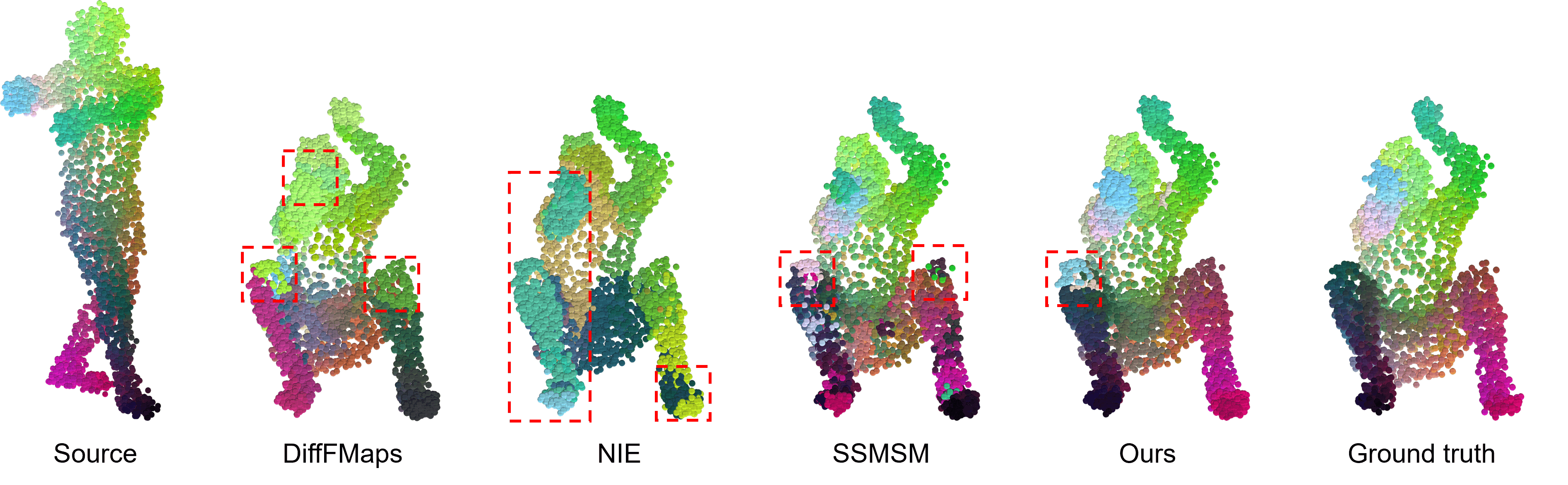}
    \caption{Robustness against additive noise. Ours produces stable correspondences under this noise compared to the baselines (errors highlighted in red).}
    \vspace*{6in}
    \label{fig:noisy}
\end{figure*}

%% file: tables/table_ablation.tex
\begin{table}
    \footnotesize
    \centering
        \begin{tabular}{@{}lccc@{}}
        \toprule
        \multicolumn{1}{c}{\multirow{1}{*}{\textbf{Geo. error ($\times$100)}}}      & \multicolumn{1}{c}{\textbf{FAUST}}   & \multicolumn{1}{c}{\textbf{SCAPE}}  \\
        \midrule
        \multicolumn{3}{c}{Ablation study on loss terms} \\
        \multicolumn{1}{l}{w/o $L_{\mathrm{off}}, L_{\mathrm{o}}$} & \multicolumn{1}{c}{20.1} & \multicolumn{1}{c}{25.8} \\
        \multicolumn{1}{l}{w/o $L_{\mathrm{o}}$} & \multicolumn{1}{c}{11.2} & \multicolumn{1}{c}{15.6} \\        
        \multicolumn{1}{l}{w/o $L_{\mathrm{c}}$} & \multicolumn{1}{c}{5.6} & \multicolumn{1}{c}{4.7} \\
        \midrule
        \multicolumn{3}{c}{Ablation study on network components} \\
        \multicolumn{1}{l}{w/o ASAP} & \multicolumn{1}{c}{3.9} & \multicolumn{1}{c}{4.3} \\
        \multicolumn{1}{l}{w/o Cross Attention} & \multicolumn{1}{c}{4.0} & \multicolumn{1}{c}{3.4} \\        
        \multicolumn{1}{l}{w/o ASAP, Cross Attention} & \multicolumn{1}{c}{4.1} & \multicolumn{1}{c}{3.6} \\
        \midrule
        \multicolumn{3}{c}{Ablation study on shape communication} \\
        \multicolumn{1}{l}{w/o Cross Attention, $L_{\mathrm{c}}$} & \multicolumn{1}{c}{4.6} & \multicolumn{1}{c}{4.9} \\
        \multicolumn{1}{l}{Ours} & \multicolumn{1}{c}{\textbf{3.7}} & \multicolumn{1}{c}{\textbf{3.2}} \\        
        \bottomrule
        \end{tabular} 
    \caption{Ablation study of our loss and pipeline. Each loss term and network component contributes to reduce matching errors.
    }
    \label{tab:ablation}
\end{table}